\documentclass{article}

\usepackage[nonatbib,final]{neurips_2022}

\usepackage[utf8]{inputenc} 
\usepackage[T1]{fontenc}    
\usepackage{hyperref}       
\usepackage{url}            
\usepackage{booktabs}       
\usepackage{amsfonts}       
\usepackage{nicefrac}       
\usepackage{microtype}      
\usepackage{xcolor}         
\usepackage{bm}
\usepackage{graphicx}  
\usepackage{enumerate}
\usepackage{multirow}
\usepackage{algorithmic}
\usepackage{arydshln}
\usepackage{subfigure}
\usepackage{enumitem}
\usepackage{physics}
\usepackage{algorithm}
\usepackage{threeparttable}
\usepackage{caption}
\newtheorem{theorem}{Theorem}
\newtheorem{lemma}{Lemma}

\newcommand{\proposed}[0]{\textsc{Smile}}
\newcommand{\proposedle}[0]{\textsc{Smile-si}}
\newcommand{\role}[0]{\textsc{Role}} 
\newcommand{\an}[0]{\textsc{An}} 
\newcommand{\anls}[0]{\textsc{An-ls}} 
\newcommand{\glocal}[0]{\textsc{Glocal}} 
\newcommand{\mlml}[0]{\textsc{Mlml}} 
 
\newcommand{\dml}[0]{\textsc{D2ml}} 
\newcommand{\wan}[0]{\textsc{Wan}}
\title{One Positive Label is Sufficient:\\Single-Positive Multi-Label	Learning  with Label Enhancement}

\author{%
	Ning~Xu$^1$, Congyu~Qiao$^1$, Jiaqi Lv$^2$, Xin Geng$^1$\thanks{Corresponding author},  Min-Ling Zhang$^1$ \\
	$^1$School of Computer Science and Engineering, Southeast University, Nanjing 210096, China\\
	$^2$ RIKEN Center for Advanced Intelligence Project, Tokyo 103-0027, Japan\\
	\texttt{\{xning, qiaocy\}@seu.edu.cn}, 	\texttt{is.jiaqi.lv@gmail.com}, \\	\texttt{\{xgeng, zhangml\}@seu.edu.cn}
}

\begin{document}

\maketitle

\begin{abstract}
Multi-label learning (MLL) learns from the examples each associated with multiple labels simultaneously, where the high cost of annotating all relevant labels for each training example is challenging for real-world applications. To cope with the challenge, we investigate single-positive multi-label learning (SPMLL) where each example is annotated with only one relevant label and show that one can successfully learn a theoretically grounded multi-label classifier on SPMLL training examples. In this paper,  a novel  SPMLL method named {\proposed}, i.e., Single-positive MultI-label	learning  with Label Enhancement, is proposed. Specifically, an unbiased risk estimator is derived, which could be guaranteed to approximately converge to the optimal risk minimizer in fully supervised learning and shows that one positive label of each instance is sufficient to train a  model. Then, the corresponding empirical risk estimator is established via recovering the latent soft label as a label enhancement process, where the posterior density of the latent soft labels is approximate to the variational Beta density parameterized by an inference model. Experiments on twelve corrupted MLL datasets show  the effectiveness of {\proposed} over several existing SPMLL approaches.  Source code is available at \url{https://github.com/palm-ml/smile}.
\end{abstract}

\section{Introduction}
Multi-label learning (MLL) aims to build a predictive model to assign  a set of relevant labels for the  unseen instance  via learning from the training examples  associated with multiple class labels simultaneously \cite{tsoumakas2006multi,zhang2014review}.  During the past decade, MLL has been widely applied to learn from the data containing rich semantics,  such as multimedia
content annotation \cite{you2020cross,vallet2015multi}, text categorization \cite{tang2020multi,rubin2012statistical}, music
emotion analysis \cite{lo2011cost,wu2014music}, and bioinformatics analysis \cite{chen2016deep}, etc.  

\begin{figure}[t]
	\centering
	\includegraphics[width=0.40\textwidth]{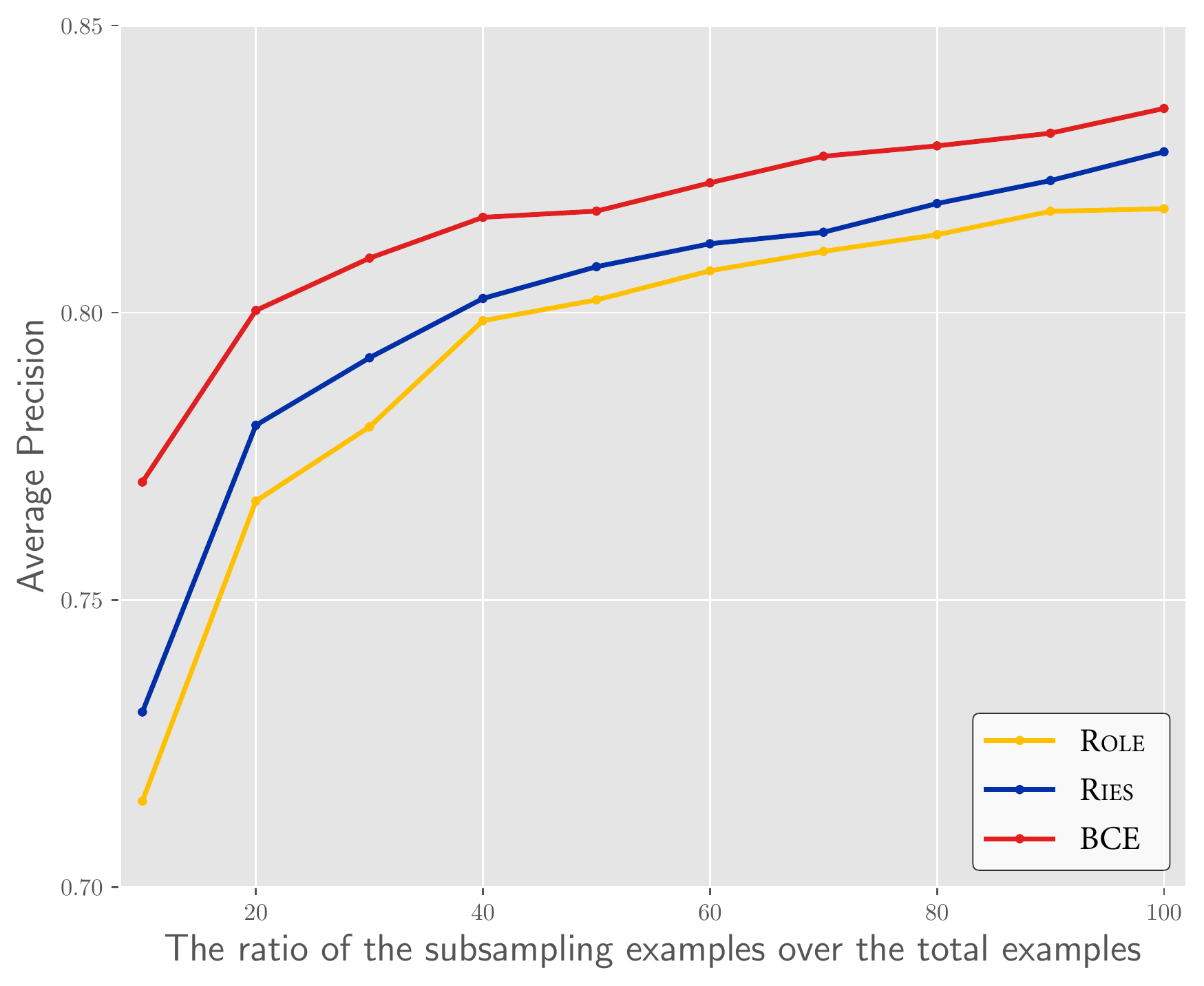}
	\caption{Test average precision  on \texttt{tmc2007}. Each curve is generated by randomly subsampling  the examples from the training set, where BCE is trained on fully labeled examples via binary cross-entropy loss  while the SPMLL methods ({\role} \cite{cole2021multi} and the proposed {\proposed}) are trained on single-positive case.}\label{SPMLL}
\end{figure}

However, in practice, obtaining  ground-truth multiple labels for MLL training datasets is  costly due to the expensive and time-consuming manual annotations. 
Comparing with multi-class learning where an example is associated with only \emph{one positive label}, multi-label learning requires the \emph{complete positive  label set} for each example. On this account, the annotation cost of multi-label learning is  significantly higher than multi-class classification, which  limits its application especially when the number of categories is large.

To mitigate this problem, the setting of single-positive multi-label learning (SPMLL) \cite{cole2021multi}  allows for significantly reduced annotations costs for the datasets,  where each example is annotated with only one relevant label.   In Figure \ref{SPMLL}, comparing with fully labeled case, the SPMLL approaches  on single-positive labeled examples  only incur a tolerable drop  in the performance but  drastically reduce the amount of supervision required to train multi-label classifiers. By establishing the SPMLL methods with the learning power of DNNs,   recent work \cite{cole2021multi} has also empirically validated that SPMLL  would  reduce the annotations costs  while achieving good performance in practice.   However,  no method  could provide theoretical insights as to why the model trained on the SPMLL examples can converge to an ideal one. 

In this paper, we propose a theoretically-guaranteed method named {\proposed}, i.e.,
Single-positive MultI-label	learning  with Label Enhancement. Specifically, we first derive an unbiased risk estimator, which  suggests that  one positive label of each instance is sufficient to train the predictive models for multi-label learning. Besides, an estimation error bound is  derived, which guarantees the risk-consistency \cite{mohri2018foundations} of the proposed method. Then we could design a benchmark solution via  recovering the soft labels   corresponding to each  example in a  label enhancement process \cite{Xu2022variational,xu2019label}, where  the posterior density of the latent soft labels is inferred by leveraging an approximate Beta density.  The contributions are summarized as follows:
\begin{itemize}[topsep=0ex,leftmargin=*,parsep=1pt,itemsep=1pt]
	\item Theoretically, we for the first time derive an unbiased risk estimator for SPMLL. 	Based on this,   an estimation error bound is established that guarantees the risk-consistency and demonstrates  that the	obtained  risk minimizer in SPMLL would approximately converge to the optimal risk minimizer in fully supervised MLL.
	\item Practically, we propose the method {\proposed} for SPMLL via adopting the latent soft labels recovered by  label enhancement. The posterior density of the latent soft label is inferred  by leveraging an approximate Beta density and   the  evidence lower bound (ELBO) \cite{kingma2014auto} for the optimization is deduced.	
\end{itemize}
Experiments on twelve corrupted MLL datasets show  the effectiveness of {\proposed} over several existing SPMLL approaches.

\section{Related Work}
In multi-label learning, each example is associated with  multiple class labels simultaneously.   As the output space in MLL is exponential in size to the number of class labels, numerous
approaches are proposed to exploit the label correlations to promote the learning process \cite{hartvigsen2020recurrent,zhang2018multi,tsoumakas2009mining}.  The first-order approaches  disassemble the MLL problem into a number of binary classification problems \cite{boutell2004learning,zhang2007ml}.  The second-order approaches consider the label correlations between pairs of  labels \cite{elisseeff2002kernel,furnkranz2008multilabel}.   The high-order approaches further focus on the label correlations among the label set \cite{read2011classifier,tsoumakas2011random}. Another line of research focuses on  manipulating the feature space via formalizing label-specific feature to each class label to facilitate multi-label classification \cite{ma2020topic,huang2016learning,yu2021multi}. In addition, some work focus on dealing with MLL via deep models.  A directed graph   over the labels  is established via  employing  GCN to propagate information among all label nodes \cite{chen2019multi}. Transformer is leveraged  for exploring the label dependency by introducing a
ternary encoding scheme to represent the state of labels \cite{lanchantin2021general}.

In practice, the labeling information is often incomplete in training data, since acquiring exhaustive supervision  is extremely difficult. Numerous approaches have been
proposed to handle the MLL with missing labels  \cite{GoldbergZRXN10}, which is also termed as  MLL with partial labels \cite{durand2019learning}. A transductive learning method is proposed to   concatenate features and labels and apply the matrix completion technique to it \cite{GoldbergZRXN10}.  Then, the  inductive learning method is proposed to exploit the structure of
specific loss functions to offer efficient algorithms for learning with missing labels  \cite{yu2014large}.  Wu \cite{wu2014multi} recovers the full	labeling information  of each  training sample via enforcing consistency with	available label-assignment and smoothness of label-assignment.  The global and local label correlations are exploited simultaneously  via learning a latent
label representation in  the missing label cases  \cite{zhu2017multi}.  Durand \cite{durand2019learning}
empirically compare different label-assignment strategies to show
the potential to employ partial labels for MLL.
Another method induces a cost function that measures the smoothness of labels and  features to alleviate the overfitting issue when
training data contains missing labels \cite{huynh2020interactive}.

Comparing with multi-label learning with missing labels, SPMLL \cite{cole2021multi}  considers the hardest version of this problem, where annotators are only asked to provide a single positive label for each training example
and no additional negative or positive labels.  When collecting multi-label annotations, it may be more efficient to annotate  only one label  rather than  multiple labels for each example. To learn from SPMLL examples, an intuitive solution is “assume negative” (AN) \cite{cole2021multi}, which assumes that unobserved labels are negative and trains the predictive model with binary cross-entropy loss on observed positive labels. Recent work \cite{cole2021multi} proposes some methods to   reduce the damaging
effects of false-negative labels. An expected positive regularization \cite{cole2021multi} is proposed to  avoid the problem but the expected number of
positive labels of each example should be given.   Label smoothing \cite{szegedy2016rethinking} is employed to  reduce  the impact of the incorrect labels.  Another approach \cite{cole2021multi}  estimates  the unobserved	labels and  encourages the  classifier predictions  to match the estimated labels via  binary entropy loss.  However,  no methods  can provide theoretical insights as to why the model trained on SPMLL examples can converge to an ideal one.  

\section{Problem Setup}
\subsection{Multi-Label Learning}
In MLL, each example  is associated with multiple labels,  and aims to build a predictive model which can assign a set of relevant labels for the unseen instance. Let $\mathcal{X} = \mathbb{R}^q$ be the $q$-dimensional instance space and $\mathcal{Y}  = \{1,2,...,c\}$ be the label space with $c$ class labels.  Given the MLL training set $ \mathcal{D} = \{(\bm{x}_i, Y_i) |1 \leq i \leq n\}$ where $\bm{x}_i \in \mathcal{X}$ denotes the $q$-dimensional instance and $Y_{i} \in \mathcal{C}$ is the set of relevant labels associated with  $\bm{x}_i$ where  $\mathcal{C}=2^{\mathcal{Y}}$.  The task of multi-label learning is to induce a multi-label classifier  $ f: \mathcal{X} \mapsto 2^\mathcal{Y} $ that minimizes the following classification risk:
\begin{equation}\label{risk_mll}
	R(f) = \mathbb{E}_{p(\bm{x}, Y)}\left[ \mathcal{L}\left( f\left(\bm{x}\right),Y\right)\right].
\end{equation}
Here,   $ \mathcal{L}: \mathbb{R}^q \times 2^{\mathcal{Y}} \mapsto \mathbb{R}_+ $ is a multi-label loss function that measures how well the model fits the data. 
Note that a method is risk-consistent if the method possesses a classification risk estimator that is equivalent to $R(f)$ given the same classifier $f$  \cite{mohri2018foundations,feng2020provably}.
\subsection{Single-Positive Multi-Label Learning}
Given the SPMLL training set $ \widetilde{\mathcal{D}} = \{(\bm{x}_i, \gamma_i) |1 \leq i \leq n\}$ where  $\gamma_{i} \in \mathcal{Y}$ denotes the observed single-positive label of $\bm{x}_i$.  Note that $\gamma_{i} \in Y_i$ while its  relevant label set $Y_i$ is not directly accessible to the   learning algorithms.  For each SPMLL training example $(\bm{x}_i, \gamma_i)$, we use the observed single-positive  vector $\bm{l}_i=[l_i^{1},l_i^{2},\ldots,l_i^{c}]^\top \in \{0,1\}^c$ to represent whether $j$-th label is the observed positive label, i.e.,  $l_i^{j} = 1$ if $j=\gamma_i $, otherwise $l_i^{j} = 0$. The multi-label vector is denoted by $\bm{y}_i=[y_i^{1},y_i^{2},\ldots,y_i^{c}]^\top \in \{0,1\}^c$, where $y_i^j=1$ if the $j$-th label is relevant to $\bm{x}_i$ and $y_i^j = 0$ if the  $j$-th label is  irrelevant. The task of SPMLL is to induce a multi-label classifier $ f: \mathcal{X} \mapsto 2^\mathcal{Y} $ from $\widetilde{\mathcal{D}}$, which can assign a set of  relevant label set for the unseen instance.  

Recent work \cite{cole2021multi}  empirically validates that SPMLL would  reduce the amount of supervision with a tolerable damage  in classification performance.   The intuitive solution AN  is  assuming that unobserved labels are negative, which leads to the drawback that  introduces some number of false negative labels. Therefore, the SOTA approaches \cite{cole2021multi}  aim to reduce the damaging effects of false-negative  labels via employing the learning power of DNNs to achieve good performance in practice. However, there is no existing method that can provide theoretical insights.

\section{The Proposed Method}
\subsection{Risk-Consistent Estimator}
To deal with single-positive multi-label learning, the classification risk $R(f)$ in Eq. (\ref{risk_mll})  could be rewritten as
\begin{equation}\label{consistent}
	\begin{split}
		&\mathbb{E}_{p(\bm{x}, Y)}\left[ \mathcal{L}\left( f\left(\bm{x}\right),Y\right)\right] \\
		=&\int_{\bm{x}}\sum_{Y\in \mathcal{C}}	\mathcal{L}\left( f\left(\bm{x}\right),Y\right) p(Y|\bm{x})p(\bm{x})\dd{\bm{x}}\\
		=&\int_{\bm{x}}\sum_{\gamma \in\mathcal{Y}} \sum_{Y\in \mathcal{C}}	\mathcal{L}\left( f\left(\bm{x}\right),Y\right) \frac{p(Y|\bm{x})}{p(y^{\gamma}=1 |\bm{x})c}
		p(y^\gamma=1 |\bm{x})p(\bm{x})\dd{\bm{x}}\\
		=&\mathbb{E}_{p(\bm{x}, \gamma)}\bigg [ \frac{1}{p(y^\gamma=1 |\bm{x})c} \sum_{Y\in \mathcal{C}} \mathcal{L}\left( f\left(\bm{x}\right),Y\right)p(Y|\bm{x})\bigg] \\=& R_{sp}(f).\\
	\end{split}
\end{equation}

Additionally, we employ the widely used loss function in multi-label learning, i.e, binary cross-entropy loss,  as the  loss function $	\mathcal{L}\left( f\left(\bm{x}\right),Y\right) $:
\begin{equation}
	\begin{split}
		\mathcal{L}\left( f\left(\bm{x}\right),Y\right) = &\sum_{j\in Y} \log f_j(\bm{x}) + \sum_{j\notin Y} \log \left( 1 - f_j(\bm{x}) \right)\\
		=&\sum_{j\in Y}\ell^j + \sum_{j \notin Y}\bar{\ell}^j,
	\end{split}
\end{equation}
where $\ell^{j}= \log f_j(\bm{x}) $ and $\bar{\ell}^{j}= 1 - f_j(\bm{x})$. Then, $\sum_{Y\in \mathcal{C}} \mathcal{L}\left( f\left(\bm{x}\right),Y\right)p(Y|\bm{x})$  in Eq. (\ref{consistent}) could be calculated as\footnote[1]{ The detail is provided in  Appendix A.}
\begin{equation}\label{L_split}
	\begin{aligned}
		&\sum_{Y\in \mathcal{C}} \mathcal{L}\left( f\left(\bm{x}\right),Y\right)p(Y|\bm{x})
		= \sum_{j=1}^{c}d^j\ell^j + \left(1-d^j\right)\bar{\ell}^j.
	\end{aligned}
\end{equation}
Here, $d^j=p(y^j=1|\bm{x}) \in [0,1]$ would be regarded as the soft label corresponding to class $j$ for $\bm{x}$.
By substituting Eq. (\ref{L_split}) into Eq. (\ref{consistent}), we obtain the following  risk-consistent estimator for SPMLL
\begin{equation}\label{risk}
	\begin{split}
		R_{sp}(f) &= \mathbb{E}_{p(\bm{x}, \gamma)}\bigg [ \frac{1}{p(y^{\gamma}=1 |\bm{x})c}   \sum_{j=1}^{c}d^j\ell^{j} 	+ (1-d^j)\bar{\ell}^j\bigg].		
	\end{split}
\end{equation}
Therefore, we could express   the empirical risk estimator  via
\begin{equation}\label{empirical}
	\begin{split}
		\widehat{R}_{sp}(f) &= \frac{1}{n}\sum_{i=1}^{n}\bigg ( \frac{1}{p(y^{\gamma_i}=1 |\bm{x}_i)c}   \sum_{j=1}^{c} d_i^j\ell_i^{j} 	+ \left(1-d_i^j\right)\bar{\ell}_i^j\bigg).			
	\end{split}
\end{equation}
Then, we could design a benchmark solution via applying the sigmoid function on $f_{\gamma_i}(\bm{x}_i)$ to approximate $p(y^{\gamma_i}=1 |\bm{x}_i)$ and  recovering the soft label  $d_i^j$  corresponding to each  example via the label enhancement process in the following subsection.

\subsection{Training with Label Enhancement}\label{label_enhancement}

To recover the soft label vector $\bm{d}_i=[d_i^1,d_i^2,\ldots,d_i^c]^\top \in [0,1]^c$,  {\proposed}  considers  the  topological information of the feature space and estimates  adjacency matrix ${\bf A}=[a_{ij}]_{n\times n}$  with 
\begin{equation}\label{ajacency}
	a_{i j}=\left\{\begin{array}{cl}
		1 & \text { if } \bm{x}_i \in \mathcal{N}(\bm{x}_j) \\
		0 &  \text{otherwise}
	\end{array}\right.,
\end{equation} 
where $\mathcal{N}(\bm{x}_j)$ is the set for  $k$-nearest neighbors of $\bm{x}_j$. 

We assume that the latent soft label matrix $\mathbf{D} =  [\bm{d}_1,\bm{d}_2,\ldots,\bm{d}_n]$   generates the observed logical  label matrix $\mathbf{L} =[\bm{l}_1,\bm{l}_2,\ldots,\bm{l}_n]$ and the adjacency matrix ${\bf A}$. Besides, the observed instance matrix $\mathbf{X}=[\bm{x}_1,\bm{x}_2,\ldots,\bm{x}_n]$ is generated from   $\mathbf{D}$ and the latent feature matrix $\mathbf{Z}= [\bm{z}_1,\bm{z}_2,\ldots,\bm{z}_n]$.   We assume that the prior density  $ p(\bm{d})$   is a Beta density  with the minor  values $\hat{\bm{\alpha}}=[\hat{\alpha}^1,\hat{\alpha}^2,\ldots,\hat{\alpha}^c]$ and $\hat{\bm{\beta}}=[\hat{\beta}^1,\hat{\beta}^2,\ldots,\hat{\beta}^c]$, i.e., $ p\left(\bm{d}\right)= \prod_{j=1}^{c}\text{Beta}\left({d}^j \mid \hat{\alpha}^j,\hat{\beta}^j\right)$. Then the prior density  $ p(\mathbf{D})$ could be  the product of each $p\left(\bm{d}\right)$. In addition, 	We assume that the prior density $p(\bm{z})$ is a standard Gaussian and prior density $p(\mathbf{Z})$ can be represented as the product of each Gaussian $	p(\mathbf{Z}) = \prod_{i=1}^{n} Gau(\bm{z}_i | \bm{0},\bm{1})$. Then, the posterior density $p(\mathbf{D}, \mathbf{Z}|\mathbf{L},\mathbf{X},\mathbf{A})$ can be decomposed as follows:
	\begin{equation}\label{encoder}
	p(\mathbf{D},\mathbf{Z} | \mathbf{L},\mathbf{X},\mathbf{A}) = p(\mathbf{D} | \mathbf{L},\mathbf{X},\mathbf{A})p(\mathbf{Z} | \mathbf{D},\mathbf{L},\mathbf{X},\mathbf{A})
	= p(\mathbf{D} | \mathbf{L},\mathbf{X},\mathbf{A})p(\mathbf{Z} | \mathbf{D},\mathbf{X})
\end{equation}

where $\mathbf{L}$ and $\mathbf{A}$ can be removed from the condition of $p(\mathbf{Z} | \mathbf{D},\mathbf{L},\mathbf{X},\mathbf{A})$ because of the independence between $\mathbf{Z}$ and $\mathbf{L},\mathbf{A}$ when latent variable $\mathbf{D}$ is given in the condition. Here we employ  $q(\mathbf{D}|\mathbf{L},\mathbf{X},\mathbf{A})$ and $q(\mathbf{Z} | \mathbf{D},\mathbf{X})$  to approximate the true posterior $p(\mathbf{D} | \mathbf{L},\mathbf{X},\mathbf{A})$ and $p(\mathbf{Z} | \mathbf{D},\mathbf{X})$ respectively.  The  approximate posterior  $q(\mathbf{D}|\mathbf{L},\mathbf{X},\mathbf{A})$ could be the product of  Beta parameterized by   $\bm{\alpha}_i=[\alpha_i^1,\alpha_i^2,\ldots,\alpha_i^c]^\top$ and $\bm{\beta}_i=[\beta_i^1,\beta_i^2,\ldots,\beta_i^c]^\top$:
\begin{equation}\label{posterior}
	\begin{split}
		q_{\bm{w}_1}(\mathbf{D} \mid \mathbf{L}, \mathbf{X}, \mathbf{A})=\prod_{i=1}^{n}\prod_{j=1}^{c}\text{Beta}\left({d}_i^j |{\alpha}_{i}^j,\beta_i^j\right).
	\end{split}
\end{equation}
Here,  the  parameters  $ \mathbf{\Delta} = [\bm{\alpha}_1,\bm{\alpha}_2,\ldots,\bm{\alpha}_n]$ and $ \mathbf{\Phi} = [\bm{\beta}_1,\bm{\beta}_2,\ldots,\bm{\beta}_n]$ are the outputs of the inference model    parameterized by $\bm{w}_1$  as a GCN \cite{kipf2016variational} with adjacency matrix by  $\mathbf{A}$.
 Let $q_{\bm{w}_2}(\mathbf{Z} | \mathbf{D},\mathbf{X})$ be the product of  Gaussian parameterized by the mean vector $\bm{\mu}_i$ and standard deviation vector $\bm{\sigma}_i$:
	\begin{equation}\label{posterior_Z}
	q_{\bm{w}_2}(\mathbf{Z} | \mathbf{D},\mathbf{X}) = \prod_{i=1}^{n} Gau(\bm{z}_i | \bm{\mu}_i,\bm{\sigma}_i),
\end{equation}
 The parameters $\mathbf{\Lambda} = [\bm{\mu}_1,\bm{\mu}_2,\ldots,\bm{\mu}_n,\bm{\sigma}_1,\bm{\sigma}_2,\ldots,\bm{\sigma}_n]$ are the outputs of the inference model with a MLP parameterized by $\bm{w}_2$.

We derive the  evidence lower bound (ELBO) \cite{kingma2014auto} on the marginal likelihood of the model to ensure that $q_{\bm{w}}(\mathbf{D,Z}|\mathbf{L},\mathbf{X},\mathbf{A})$ is as close as possible to $p(\mathbf{D,Z}|\mathbf{L},\mathbf{X},\mathbf{A})$ \footnote[1]{ The detail is provided in  Appendix B.}:
\begin{equation}\label{lowerbound}
	\begin{split}
		\mathcal{L}_{ELBO} &= \mathbb{E}_{q_{\bm{w}}({\bf D,Z}|{\bf L}, {\mathbf{X}}, {\bf A})}[ \log p({\mathbf{X}}|{\bf D,Z})+\log p({\bf L}|{\bf D})+\log p( {\bf A}|{\bf D})]\\&- \text{KL}[ q_{\bm{w}_1}({\mathbf D}|{\mathbf L}, {\mathbf{X}}, {\mathbf A})|| p({\mathbf D})]
	- \text{KL}[q_{\bm{w}_2}(\mathbf{Z} | \mathbf{D},\mathbf{X}) || p(\mathbf{Z})].
	\end{split}
\end{equation} 

 We further assume  that $p({\mathbf{X}}|{\bf D,Z})$ is  a product of each Gaussian with means $\bm{\xi}_i$ and $p\left(\bf L| \mathbf{D}\right)$ is a product of each multivariate Bernoulli with probabilities $\bm{\tau}_i$.  In order to simplify the observation model,  $\mathbf{T}^{(m)}=[\bm{\tau}_1^{(m)}, \bm{\tau}_2^{(m)},\ldots,\bm{\tau}_n^{(m)}]$ is computed from $m$-th sampling $\mathbf{D}^{(m)}$ with a  MLP parameterized by $\bm{\eta}_1$ and  $\mathbf{\Xi}^{(m)}=[\bm{\xi}_1^{(m)}, \bm{\xi}_2^{(m)},\ldots,\bm{\xi}_n^{(m)}]$ is computed from $m$-th sampling $\mathbf{D}^{(m)}$ and $\mathbf{Z}^{(m)}$ with a  MLP parameterized by $\bm{\eta}_2$ .  Then the first part of Eq. (\ref{lowerbound}) can be tractable as
\begin{equation}\label{part1}
	\begin{split}
		\mathbb{E}_{q_{\bm{w}}({\bf D,Z}|{\bf L}, {\mathbf{X}}, {\bf A})}[\log p({\mathbf{X}}|{\bf D,Z})+\log p({\bf L}|{\bf D})+\log p( {\bf A}|{\bf D})]
		=\frac{1}{M}\sum_{m=1}^{M}  \operatorname{tr}\left( \mathbf{L}^{\top} \log \mathbf{T}^{(m)} \right) \\ 
		 +\operatorname{tr}\left( \left(\mathbf{I}-\mathbf{L}\right)^\top \log \left(\mathbf{I}-\mathbf{T}^{(m)}\right) \right) - \Vert \mathbf{A}-S\left(\mathbf{D}^{(m)}\mathbf{D}^{(m)\top}\right) \Vert_F^2 
		 + \Vert \bm{\Xi}^{(m)} - {\bf X} \Vert_F^2
	\end{split}
\end{equation}
Here,  $S(\cdot)$ is the logistic sigmoid function, and  implicit reparameterization trick \cite{figurnov2018implicit} and MC sampling \cite{kingma2014auto, xu2021instance, xu2020variational} are employed.

The second part of Eq. (\ref{lowerbound})  can be  analytically calculated as
\begin{equation}\label{D_KL}
	\begin{split}
		\operatorname{KL}\left(q_{\bm{w}_1}(\mathbf{D}|\mathbf{L},\mathbf{X},\mathbf{A}) \| p(\mathbf{D})\right)
		=& \sum\limits_{i=1}^{n}\sum\limits_{j=1}^{c}\log \frac{\Gamma(\alpha_{i}^j+\beta_i^j)\Gamma(\hat{\alpha}_{i}^j)\Gamma(\hat{\beta}_i^j) }{\Gamma(\hat{\alpha}_{i}^j+\hat{\beta}_i^j)\Gamma({\alpha}_{i}^j)\Gamma({\beta}_i^j) }
		+(\alpha_i^j-\hat{\alpha}_i^j)\psi(\alpha_i^j)
		\\-&(\alpha_i^j-\hat{\alpha}_i^j+\beta_i^j-\hat{\beta}_i^j)\psi(\alpha_i^j+\beta_i^j)
		+(\beta_i^j-\hat{\beta}_i^j)\psi(\beta_i^j).		
	\end{split}
\end{equation}
Here,  $\Gamma(\cdot)$ and $\psi(\cdot)$ are  Gamma function and Digamma function, respectively. 
The third  part of Eq. (\ref{lowerbound})  can be analytically calculated as follows:
\begin{equation}\label{Z_KL}
	\begin{split}
		\text{KL}(q_{\bm{w}_2}(\mathbf{Z} | \mathbf{D},\mathbf{X}) || p(\mathbf{Z}))=\sum_{i=1}^{n}\sum_{j=1}^J \bigg (1+\log((\sigma_i^j))
		-(\mu_i^j)^2-(\sigma_i^j)^2 \bigg ).
	\end{split}
\end{equation}

Besides, we could  promote the label enhancement process  via enforcing that the estimated $\bf D$ should inherit the labeling-information of observed labels:
\begin{equation}
	\begin{split}
		{T}_{C}=-\frac{1}{n} \sum_{i=1}^{n} \sum_{j=1}^{c} l_i^j\log d_{i}^{j} + \left(1-l_i^j\right)\left(1-\log d_{i}^{j}\right).
	\end{split}
\end{equation}

Finally,  the  objective  of label enhancement ${T}_{LE}$ is obtained: 
\begin{equation}
	\label{enhancement}
	\begin{split}
		{T}_{LE}=- \lambda\mathcal{L}_{ELBO}+{T}_{C},
	\end{split}
\end{equation}
where $\lambda$ is a hyper-parameter.

{\proposed} first initializes the predictive network by warm-up training with AN solution, which would attain a fine  network before it starts fitting noise \cite{ZhangBHRV17}.
Then  we could sample the  soft label  from fixed Beta after label enhancement and the sigmoid function on $f_{\gamma_i}(\bm{x}_i)$ to approximate $p(y^{\gamma_i}=1 |\bm{x}_i)$ to make Eq. (\ref{empirical}) accessible, and    train  the predictive model $\bm{\theta}$ by minimizing the  risk estimator.   In each epoch,   {\proposed} alternately operates label enhancement process and classifier training process.   Algorithm \ref{alg:Framwork} shows the algorithmic description of  {\proposed}.

\begin{algorithm}[t] 
	\caption{ {\proposed} Algorithm} 
	\label{alg:Framwork} 
	\begin{algorithmic}[1] 
		\REQUIRE The SPMLL training set $\widetilde{\mathcal{D}}= \{(\bm{x}_i,\gamma_i)\}_{i=1}^n$, the number of iteration $I$ and the number of epoch $T$;
		\STATE Warm-up  $\bm{\theta}$ by using AN solution, and initialize the reference model $\bm{w}_1$,  $\bm{w}_2$ and observation model $\bm{\eta}$;
		\STATE Estimate the adjacency matrix $\mathbf{A}$ by Eq. (\ref{ajacency});
		\FOR {$t=1,\ldots, T$}
		\STATE Shuffle training set $\widetilde{\mathcal{D}} = \{(\bm{x}_i,\gamma_i)\}_{i=1}^n$ into $I$ mini-batches;
		\FOR {$k=1,\ldots, I$}
		\STATE Update $\bm{w}_1$,  $\bm{w}_2$ and $\bm{\eta}$ by forward computation and back-propagation by  Eq. (\ref{enhancement}); 
		\STATE Obtain the soft label    $\bm{d}_i$ for each example $\bm{x}_i$ by Eq. (\ref{posterior}); 
		\STATE Apply the sigmoid function on $f_{\gamma_i}(\bm{x}_i)$ to approximate $p(y^{\gamma_i}=1 |\bm{x}_i)$; 
		\STATE Update  $\bm{\theta}$ by forward computation and back-propagation by  Eq. (\ref{empirical}); 
		\ENDFOR
		\ENDFOR
		\ENSURE The predictive model $\bm{\theta}$.
	\end{algorithmic} 
\end{algorithm}
\subsection{Estimation Error Bound}

In this subsection, we establish an estimation error bound of the proposed method. The empirical risk estimator according to Eq.(\ref{empirical}) can be rewritten as:
\begin{equation}
	\widehat{R}_{sp}(f) =\frac{1}{n}\sum_{i=1}^{n}\sum_{j=1}^{L}\left({w}_i^j\ell_i^j + \bar{w}_i^j\bar{\ell}_i^j\right), 
\end{equation}
where ${w}_i^j = \frac{d_i^j}{p(y^\gamma=1|\bm{x}_i)c}$ and $\bar{w}_i^j =\frac{1-d_i^j}{p(y^\gamma=1|\bm{x}_i)c}$. Then the loss function $\mathcal{L}_{sp}$ is 
\begin{equation}
	\mathcal{L}_{sp}=\sum_{j=1}^{L}\left({w}_i^j\ell_i^j + \bar{w}_i^j\bar{\ell}_i^j\right). 
\end{equation}

We define a function space as:
\begin{equation}
	\mathcal{G}_{sp} = \left\{(\bm{x}, y) \mapsto \sum_{j=1}^{L}\left({w}^j\ell^j + \bar{w}^j\bar{\ell}^j\right) | f\in\mathcal{F} \right\},
\end{equation}
and denote  the expected Rademacher complexity \cite{bartlett2002rademacher} of $\mathcal{G}_{sp}$ as: 
\begin{equation}
	\widetilde{\mathfrak{R}}_n(\mathcal{G}_{sp}) = \mathbb{E}_{\bm{x},y,\pmb{\sigma}}\left[\sup_{g\in\mathcal{G}_{sp}}\frac{1}{n}\sum_{i=1}^{n}\sigma_ig\left( \bm{x}_i ,y_i\right) \right],
\end{equation}
where  $\pmb{\sigma}=\{\sigma_1, \sigma_2,\dots, \sigma_n\}$ is $n$ Rademacher variables with $\sigma_i$ independently uniform variable taking value in $\{+1, -1\}$.
Then we have
\begin{lemma} \label{lemma_1}
	We suppose that the SPMLL loss function $\mathcal{L}_{sp}$ could be bounded by $M$, i.e., $M =\sup_{\bm{x}\in \mathcal{X}, f\in\mathcal{F}, y\in\mathcal{Y}}\mathcal{L}_{sp}\left(f(\bm{x}), y\right)$, and  for any $\delta \textgreater 0$, with probability at least $1 - \delta$, then we have
	$$
	\sup_{f\in\mathcal{F}}\abs{R_{sp}(f)-\widehat{R}_{sp}(f)}\leq 2\widetilde{\mathfrak{R}}_n(\mathcal{G}_{sp})
	+ \frac{M}{2} \sqrt{\frac{\log\frac{2}{\delta}}{2n}}.
	$$
\end{lemma}
The proof of Lemma \ref{lemma_1} could be founded  in Appendix C.
\begin{lemma}\label{lemma_2}
	We suppose that  the loss function $\ell\left(f(\bm{x}),{y}\right)$ and $\bar{\ell}\left(f(\bm{x}),{y}\right)$ are $\rho^{+}$-Lipschitz and $\rho^{-}$-Lipschitz with respect to $f(\bm{x})$ $(0\textless \rho^{+} \textless \infty$ and $0\textless \rho^{-} \textless \infty )$ for all $y \in \mathcal{Y}$, respectively,  and $w^j$ and $\bar{w}^j$  are both  bounded in $[0, \kappa]$. Then, we have 
	\begin{equation}
		\widetilde{\mathfrak{R}}_n(\mathcal{G}_{sp}) \leq \sqrt{2}\kappa c(\rho^{+} + \rho^{-})\sum_{j=1}^{c}\mathfrak{R}_n(\mathcal{H}_{y_j}),
		\nonumber
	\end{equation}
	where			$\mathcal{H}_{y} = \{ h: \bm{x} \mapsto f_{y}(\bm{x})|f\in \mathcal{F}\}$ and
	$	\mathfrak{R}_n(\mathcal{H}_{y}) = \mathbb{E}_{\bm{x},\pmb{\sigma}}\left[\sup_{h\in\mathcal{H}_{y}}\frac{1}{n}\sum_{i=1}^{n}h\left( \bm{x}_i\right) \right]$.

\end{lemma}
The proof of Lemma \ref{lemma_2}  could be founded  in Appendix D.

\begin{table*}[tb]
	\centering
	\fontsize{6}{7}\selectfont
	\caption{Predictive performance of each comparing approach (mean$\pm$std) in terms of  \emph{Average precision} $\uparrow$.  The best performance (the larger the better) is shown in bold face.}
	\begin{tabular}{ccccccccc}
		\toprule
		Datasets &{\proposed} &{\an} & {\anls} & {\wan} & {\role} & {\glocal} & {\mlml} & {\dml}  \\
		\midrule
		CAL500 & \textbf{0.401$\pm$0.011} & 0.382$\pm$0.044 & 0.253$\pm$0.031 & 0.393$\pm$0.011 & 0.288$\pm$0.008 & 0.227$\pm$0.002 & 0.233$\pm$0.000 & 0.223$\pm$0.001 \\ 
		image & \textbf{0.784$\pm$0.044} & 0.613$\pm$0.081 & 0.621$\pm$0.073 & 0.685$\pm$0.058 & 0.696$\pm$0.039 & 0.771$\pm$0.003 & 0.652$\pm$0.001 & 0.274$\pm$0.003 \\ 
		scene & \textbf{0.841$\pm$0.070} & 0.740$\pm$0.127 & 0.741$\pm$0.117 & 0.801$\pm$0.020 & 0.717$\pm$0.067 & 0.825$\pm$0.001 & 0.814$\pm$0.000 & 0.285$\pm$0.002 \\ 
		yeast & \textbf{0.758$\pm$0.003} & 0.755$\pm$0.003 & 0.753$\pm$0.003 & 0.757$\pm$0.003 & 0.753$\pm$0.003 & 0.646$\pm$0.002 & 0.456$\pm$0.002 & 0.323$\pm$0.001 \\ 
		corel5k & \textbf{0.303$\pm$0.007} & 0.299$\pm$0.005 & 0.272$\pm$0.005 & 0.302$\pm$0.004 & 0.215$\pm$0.011 & 0.218$\pm$0.001 & 0.072$\pm$0.001 & 0.028$\pm$0.001 \\ 
		rcv1-s1 & \textbf{0.616$\pm$0.001} & 0.604$\pm$0.004 & 0.581$\pm$0.002 & 0.610$\pm$0.005 & 0.570$\pm$0.004 & 0.229$\pm$0.000 & 0.221$\pm$0.003 & 0.053$\pm$0.001 \\ 
		corel16k-s1 & \textbf{0.344$\pm$0.003} & 0.337$\pm$0.003 & 0.316$\pm$0.002 & 0.344$\pm$0.003 & 0.288$\pm$0.004 & 0.029$\pm$0.001 & 0.081$\pm$0.001 & 0.029$\pm$0.004 \\ 
		delicious & 0.319$\pm$0.001 & 0.297$\pm$0.009 & 0.193$\pm$0.005 & \textbf{0.320$\pm$0.001} & 0.199$\pm$0.004 & 0.027$\pm$0.001 & 0.086$\pm$0.001 & 0.028$\pm$0.001 \\ 
		iaprtc12 & \textbf{0.314$\pm$0.003} & 0.292$\pm$0.008 & 0.244$\pm$0.008 & 0.266$\pm$0.006 & 0.243$\pm$0.005 & 0.035$\pm$0.002 & 0.126$\pm$0.001 & 0.026$\pm$0.001 \\ 
		espgame & \textbf{0.259$\pm$0.003} & 0.248$\pm$0.002 & 0.208$\pm$0.003 & 0.259$\pm$0.002 & 0.216$\pm$0.004 & 0.038$\pm$0.000 & 0.086$\pm$0.002 & 0.038$\pm$0.001 \\ 
		mirflickr & \textbf{0.635$\pm$0.004} & 0.629$\pm$0.003 & 0.604$\pm$0.004 & 0.611$\pm$0.004 & 0.545$\pm$0.019 & 0.476$\pm$0.000 & 0.253$\pm$0.003 & 0.132$\pm$0.002 \\ 
		tmc2007 & \textbf{0.820$\pm$0.002} & 0.815$\pm$0.003 & 0.802$\pm$0.003 & 0.815$\pm$0.001 & 0.798$\pm$0.005 & 0.649$\pm$0.000 & 0.415$\pm$0.000 & 0.161$\pm$0.001 \\ 
		\bottomrule
	\end{tabular}
	\label{average}
\end{table*}

Based one Lemma \ref{lemma_1} and \ref{lemma_2}, we could  obtain the following theorem:
\begin{theorem}\label{theorem}
	Assume the loss function $\ell\left(f(\bm{x}),{y}\right)$ and $\bar{\ell}\left(f(\bm{x}),{y}\right)$ are $\rho^{+}$-Lipschitz and $\rho^{-}$-Lipschitz with respect to $f(\bm{x})$ $(0\textless \rho^{+} \textless \infty$ and $0\textless \rho^{-} \textless \infty )$ for all $y \in \mathcal{Y}$ and the loss function $\mathcal{L}_{sp}$ are bounded by $M$, i.e., $M =\sup_{\bm{x}\in \mathcal{X}, f\in\mathcal{F}, y\in\mathcal{Y}}\mathcal{L}_{sp}\left(f(\bm{x}), y\right)$, with probability at least $1 - \delta$,
	\begin{equation}
		\begin{aligned}
			R(\widehat{f}_{sp}) - R(f^{*})&\leq 4\sqrt{2}\kappa c(\rho^{+} + \rho^{-})\sum_{j=1}^{c}\mathfrak{R}_n(\mathcal{H}_{y}) 
			+ M \sqrt{\frac{\log\frac{2}{\delta}}{2n}}.
		\end{aligned}
		\nonumber
	\end{equation}
\end{theorem}
Here,  $\widehat{f}_{\mathrm{sp}}=\min _{f \in \mathcal{F}} \widehat{R}_{\mathrm{sp}}(f)$ and  $f^{\star}=\min _{f \in \mathcal{F}} R(f)$ are the empirical risk minimizer and the true risk minimizer, respectively.  The proof could be founded  in Appendix E.   Theorem \ref{theorem} shows that  $f_{sp}$ would converge to    $f^{\star}$ as $n \rightarrow \infty$ and $\mathfrak{R}_{n}\left(\mathcal{H}_{y}\right) \rightarrow 0$.

\begin{table*}[t]
	\centering
	\fontsize{6}{7}\selectfont
	\caption{Predictive performance of each comparing approach (mean$\pm$std) in terms of  \emph{One-error} $\downarrow$.  The best performance (the smaller the better) is shown in bold face.}
	\begin{tabular}{ccccccccc}
		\toprule
		Datasets &{\proposed} &{\an} & {\anls} & {\wan} & {\role} & {\glocal} & {\mlml} & {\dml}  \\
		\midrule
		CAL500 & 0.358$\pm$0.156 & \textbf{0.325$\pm$0.134} & 0.627$\pm$0.188 & 0.420$\pm$0.152 & 0.557$\pm$0.034 & 0.843$\pm$0.011 & 0.839$\pm$0.032 & 0.833$\pm$0.003 \\ 
		image & 0.350$\pm$0.046 & 0.597$\pm$0.102 & 0.577$\pm$0.095 & 0.516$\pm$0.090 & 0.488$\pm$0.055 & 0.365$\pm$0.012 & \textbf{0.200$\pm$0.023} & 0.600$\pm$0.019 \\ 
		scene & 0.278$\pm$0.112 & 0.417$\pm$0.152 & 0.412$\pm$0.145 & 0.344$\pm$0.033 & 0.477$\pm$0.111 & 0.286$\pm$0.024 & \textbf{0.167$\pm$0.011} & 0.667$\pm$0.023 \\ 
		yeast & \textbf{0.236$\pm$0.008} & 0.242$\pm$0.012 & 0.244$\pm$0.009 & 0.242$\pm$0.011 & 0.239$\pm$0.010 & 0.276$\pm$0.032 & 0.285$\pm$0.003 & 0.500$\pm$0.022 \\ 
		corel5k & \textbf{0.648$\pm$0.008} & 0.685$\pm$0.019 & 0.674$\pm$0.013 & 0.656$\pm$0.013 & 0.696$\pm$0.022 & 0.764$\pm$0.011 & 0.947$\pm$0.005 & 0.987$\pm$0.003 \\ 
		rcv1-s1 & \textbf{0.438$\pm$0.007} & 0.445$\pm$0.011 & 0.467$\pm$0.008 & 0.449$\pm$0.011 & 0.464$\pm$0.003 & 0.810$\pm$0.029 & 0.782$\pm$0.002 & 0.941$\pm$0.004 \\ 
		corel16k-s1 & \textbf{0.641$\pm$0.008} & 0.655$\pm$0.005 & 0.666$\pm$0.007 & 0.642$\pm$0.009 & 0.667$\pm$0.009 & 0.989$\pm$0.003 & 0.830$\pm$0.001 & 0.987$\pm$0.003 \\ 
		delicious & 0.410$\pm$0.005 & 0.454$\pm$0.026 & 0.516$\pm$0.021 & \textbf{0.405$\pm$0.007} & 0.498$\pm$0.012 & 0.996$\pm$0.003 & 0.804$\pm$0.011 & 0.967$\pm$0.003 \\ 
		iaprtc12 & \textbf{0.579$\pm$0.022} & 0.604$\pm$0.022 & 0.618$\pm$0.017 & 0.662$\pm$0.015 & 0.659$\pm$0.016 & 0.997$\pm$0.000 & 0.605$\pm$0.012 & 0.897$\pm$0.006 \\ 
		espgame & \textbf{0.662$\pm$0.007} & 0.686$\pm$0.012 & 0.702$\pm$0.011 & 0.673$\pm$0.008 & 0.707$\pm$0.009 & 0.995$\pm$0.000 & 0.699$\pm$0.003 & 0.734$\pm$0.004 \\ 
		mirflickr & \textbf{0.335$\pm$0.013} & 0.343$\pm$0.017 & 0.343$\pm$0.006 & 0.385$\pm$0.009 & 0.497$\pm$0.030 & 0.670$\pm$0.054 & 0.447$\pm$0.011 & 0.816$\pm$0.007 \\ 
		tmc2007 & \textbf{0.204$\pm$0.003} & 0.215$\pm$0.003 & 0.221$\pm$0.006 & 0.220$\pm$0.003 & 0.225$\pm$0.004 & 0.313$\pm$0.001 & 0.227$\pm$0.002 & 0.409$\pm$0.008 \\ 
		\bottomrule
	\end{tabular}
	\label{oneerr}
\end{table*}

\section{Experiments}
\subsection{Experimental Configurations}

In the experiments, we adopt  twelve widely-used MLL datasets \cite{hang2021collaborative}, which cover a broad range of cases with diversified multi-label properties. To evaluate the performance of SPMLL methods, we generate the single positive training data by randomly selecting one positive label to keep for each training example in the MLL datasets. For each dataset, we run the comparing methods with 80\%/10\%/10\% train/validation/test split. The validation and test sets
are always fully labeled.  The detailed descriptions of these datasets are provided in Appendix F. Five popular multi-label metrics  \emph{Ranking loss}, \emph{Hamming loss}, \emph{One-error}, \emph{Coverage}, and \emph{Average precision} \cite{zhang2014review} are employed for performance evaluation. Furthermore, for \emph{Average precision}, the \emph{larger} the values the better the performance. While for the other four metrics, the \emph{smaller} the values the better the performance.

\begin{table*}[t]
	\centering
	\fontsize{6}{7}\selectfont
	\caption{ Summary of the Wilcoxon signed-ranks test for {\proposed} against other comparing approaches at 0.05 significance level. The $p$-values are shown in the brackets.}
	\begin{tabular}{cccccccc}
		\hline
		{\proposed} against &{\an} & {\anls} & {\wan} & {\role} & {\glocal} & {\mlml} & {\dml}  \\
		\hline
		\emph{Average precision} & \textbf{win}[0.0005] & \textbf{win}[0.0005] & \textbf{win}[0.0092] & \textbf{win}[0.0005] & \textbf{win}[0.0005] & \textbf{win}[0.0005] & \textbf{win}[0.0005] \\    
		\emph{One-error} & \textbf{win}[0.0122] & \textbf{win}[0.0005] & \textbf{win}[0.0015] & \textbf{win}[0.0005] & \textbf{win}[0.0005] & \textbf{win}[0.0342] & \textbf{win}[0.0005] \\
		\emph{Ranking loss} & \textbf{win}[0.0269] & \textbf{win}[0.0005] & \textbf{tie}[0.1533] & \textbf{win}[0.0005] & \textbf{win}[0.0005] & \textbf{win}[0.0024] & \textbf{win}[0.0005] \\
		\emph{Hamming loss} & \textbf{win}[0.0277] & \textbf{win}[0.0178] & \textbf{win}[0.0005] & \textbf{win}[0.0277] & \textbf{win}[0.0277] & \textbf{win}[0.0277] & \textbf{win}[0.0077] \\
		\emph{Coverage} & \textbf{win}[0.0425] & \textbf{win}[0.0005] & \textbf{tie}[0.1819] & \textbf{win}[0.0005] & \textbf{win}[0.0005] & \textbf{win}[0.0024] & \textbf{win}[0.0015] \\
		\hline
	\end{tabular}
	\label{Wilcoxon}
\end{table*}


\begin{table*}[t]
	\centering
	\scriptsize
	\caption{Predictive performance of {\proposed} and its variant (mean$\pm$std) in terms of  \emph{Average precision}, \emph{One-error}, and  \emph{Ranking loss}.  }
	\begin{tabular}{ccccccc}
		\hline
		\multirow{2}*{Datasets} & \multicolumn{2}{c}{\emph{Average precision}$\uparrow$} & \multicolumn{2}{c}{\emph{One-error}$\downarrow$} & \multicolumn{2}{c}{\emph{Ranking loss}$\downarrow$} \\ 
		
		\cline{2-7}
		~ & {\proposed} & {\proposedle} & {\proposed} & {\proposedle} & {\proposed} & {\proposedle} \\
		\hline
		CAL500 & 0.401$\pm$0.011 & \textbf{0.409$\pm$0.000} & 0.358$\pm$0.156 & \textbf{0.343$\pm$0.010} & \textbf{0.239$\pm$0.010} & 0.244$\pm$0.000 \\ 
		image & \textbf{0.784$\pm$0.044} & 0.547$\pm$0.022 & \textbf{0.350$\pm$0.046} & 0.675$\pm$0.035 & \textbf{0.170$\pm$0.055} & 0.428$\pm$0.005 \\
		scene & \textbf{0.841$\pm$0.070} & 0.711$\pm$0.057 & \textbf{0.278$\pm$0.112} & 0.483$\pm$0.093 & \textbf{0.086$\pm$0.045} & 0.168$\pm$0.037 \\
		yeast & \textbf{0.758$\pm$0.003} & 0.738$\pm$0.002 & \textbf{0.236$\pm$0.008} & 0.246$\pm$0.006 & \textbf{0.161$\pm$0.003} & 0.176$\pm$0.001 \\
		corel5k & \textbf{0.303$\pm$0.007} & 0.302$\pm$0.003 & \textbf{0.648$\pm$0.008} & 0.655$\pm$0.007 & 0.134$\pm$0.003 & \textbf{0.116$\pm$0.000} \\
		rcv1-s1 & \textbf{0.616$\pm$0.001} & 0.577$\pm$0.004 & \textbf{0.438$\pm$0.007} & 0.477$\pm$0.015 & \textbf{0.042$\pm$0.000} & 0.055$\pm$0.000 \\
		corel16k-s1 & \textbf{0.344$\pm$0.003} & 0.336$\pm$0.001 & \textbf{0.641$\pm$0.008} & 0.649$\pm$0.002 & \textbf{0.133$\pm$0.001} & 0.140$\pm$0.000 \\
		delicious & \textbf{0.319$\pm$0.001} & 0.293$\pm$0.004 & \textbf{0.410$\pm$0.005} & 0.434$\pm$0.023 & \textbf{0.126$\pm$0.000} & 0.146$\pm$0.002 \\
		iaprtc12 & \textbf{0.314$\pm$0.003} & 0.287$\pm$0.001 & \textbf{0.579$\pm$0.022} & 0.607$\pm$0.015 & \textbf{0.115$\pm$0.002} & 0.143$\pm$0.002 \\
		espgame & \textbf{0.259$\pm$0.003} & 0.249$\pm$0.003 & \textbf{0.662$\pm$0.007} & 0.675$\pm$0.002 & \textbf{0.158$\pm$0.001} & 0.170$\pm$0.002 \\
		mirflickr & \textbf{0.635$\pm$0.004} & 0.628$\pm$0.001 & \textbf{0.335$\pm$0.013} & 0.349$\pm$0.012 & \textbf{0.117$\pm$0.002} & 0.122$\pm$0.002 \\
		tmc2007 & \textbf{0.820$\pm$0.002} & 0.814$\pm$0.001 & \textbf{0.204$\pm$0.003} & 0.210$\pm$0.003 & 0.049$\pm$0.001 & \textbf{0.048$\pm$0.000} \\
		\bottomrule
	\end{tabular}
	\label{variants}
\end{table*}

In  this  paper, {\proposed} is compared against four well-established SPMLL approaches including 1) {\an}  \cite{cole2021multi} which  assumes unobserved labels	are negative and employs binary entropy loss for the training examples with the modified labels, 2) {\anls} \cite{cole2021multi} which  assumes unobserved labels	are negative and employs label smoothing \cite{szegedy2016rethinking} to  reduce the impact of the incorrect labels (i.e. those labels incorrectly assumed to be negative), 3) {\wan} \cite{cole2021multi} which reduces the impact of false negatives by employing the down-weight terms in the loss corresponding to negative labels, and 4) {\role} \cite{cole2021multi} which online estimates of the unobserved	labels throughout training and  encourages the  classifier predictions  to match the estimated labels via  binary entropy loss. 

As SPMLL could be regarded as the hardest version of MLL with missing labels, three  well-established approaches of MLL with missing labels (also termed as  MLL with partial labels) are adopted as the comparing approaches including 1)  {\glocal} \cite{zhu2017multi} which exploits global and local label correlations simultaneously via learning a latent
label representation in  the missing label cases, 2) {\mlml} \cite{wu2014multi} which recovers the full	label assignment for each sample by enforcing consistency with	available label assignments and smoothness of labels, and 3) {\dml} \cite{ma2021expand} which utilizes both local low-rank label structures and label discriminant information for learning from missing labels. 

For all the DNN-based approaches ({\an}, {\anls}, {\wan}, {\role} and {\proposed}),  we adopt three-layer MLP as the  predictive model for fair comparisons and use the Adam optimizer \cite{KingmaB14}.  The mini-batch size  and the number of epochs  are set  to 16 and  25, respectively. The learning rate and weight decay are selected from $\{10^{-4}, 10^{-3}, 10^{-2}\}$  with a validation set. Other hyper-parameters for all the comparing methods are also selected based on the validation. All the comparing methods run 5 trials (with 80\%/10\%/10\% train/validation/test split) on each dataset.

\subsection{Experimental Results}
Tables \ref{average} and \ref{oneerr}  report the   results of all   approaches  on \emph{Average precision} and \emph{One-error},  respectively. For each evaluation metric, ``$\uparrow$'' indicates “the smaller the better” while ``$\downarrow$'' indicates “the larger the better”.  The results on other metrics are similar and could be seen in Appendix F. In addition, Wilcoxon signed-ranks test \cite{demvsar2006statistical} is employed to  show whether  {\proposed} has a significant performance  than other comparing approaches. Table  \ref{Wilcoxon}  reports  the $p$-values  for  the  corresponding  tests and the statistical test results at 0.05 significance level.  

 Table \ref{Wilcoxon} shows that  {\proposed} achieves superior  performance  against all the comparing approaches  on all evaluation metrics (except  on \emph{Rranking loss} and \emph{Coverage} where {\proposed} achieves comparable  performance  against {\wan}). The superior	performance of {\proposed} provides a strong evidence for the	effectiveness of risk-consistent estimator  for SPMLL. Tables  \ref{average} and \ref{oneerr} show that the performance advantage of {\proposed} over comparing approaches is stable under  varying  the    number  of  class labels. In summary, these experimental results clearly validate the effectiveness of {\proposed}.

\begin{figure}[t]
	\begin{minipage}{0.3\linewidth}
		\centering{\includegraphics[width=1\textwidth]{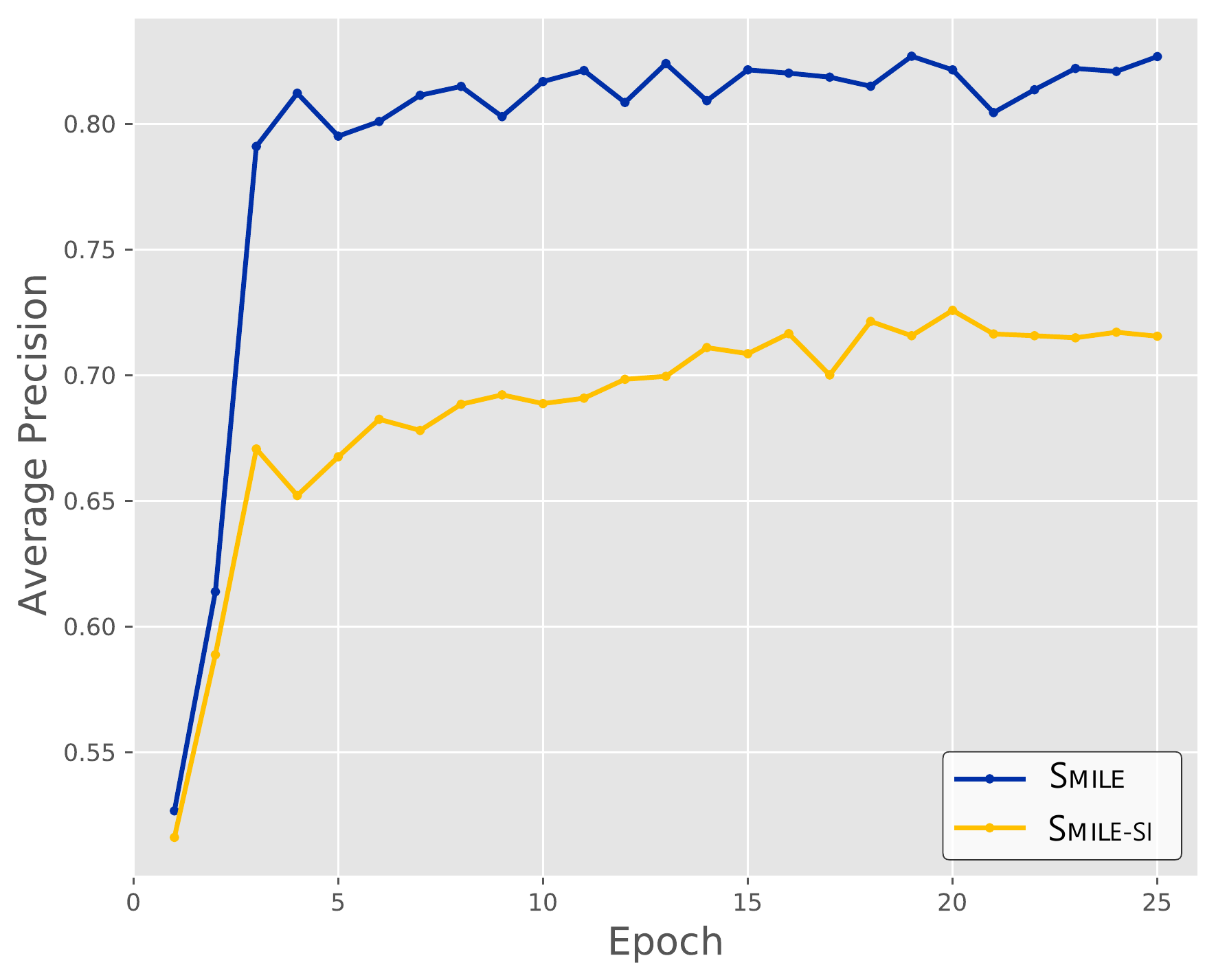}}
		\caption{\emph{Average precision}.}\label{fig:ablation}
	\end{minipage}
	\begin{minipage}{0.3\linewidth}
	\centering{\includegraphics[width=1\textwidth]{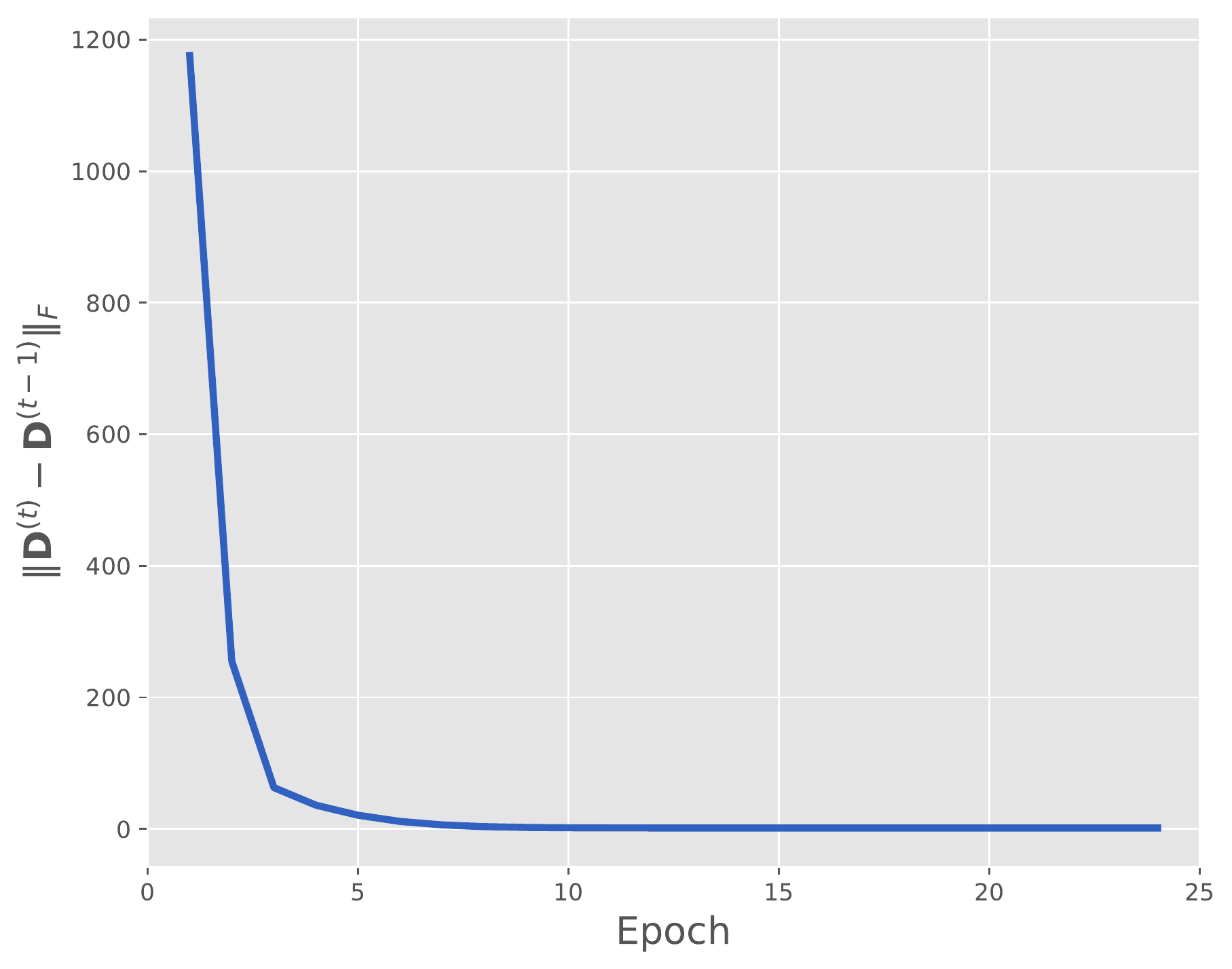}}
	\caption{Convergence  of $\mathbf{D}$.}\label{fig:convergence}
\end{minipage}
	\hfill
	\begin{minipage}{0.38\linewidth}
	\scriptsize
		\captionof{table}{Wilcoxon signed ranks test(at 0.05 significance level).}\label{Wilcoxon_variants}
	\begin{tabular}{lcc}
	\hline
	\hline
	\multirow{2}[2]{*}{Evaluation metric} & \multicolumn{2}{c}{{\proposed} against {\proposedle}} \\
	\cmidrule{2-3}          & performance & $p$-value \\
	\hline
	\emph{Average precision} & \textbf{win} & $0.0049$ \\
	\emph{One-error}         & \textbf{win} & $0.0092$ \\
	\emph{Ranking loss}      & \textbf{win} & $0.0161$ \\
	\emph{Hamming loss}      & \textbf{win} & $0.0277$ \\
	\emph{Coverage}          & \textbf{win} & $0.0122$ \\
	\hline
	\hline
\end{tabular}
	\end{minipage}
\end{figure}

\subsection{Further Analysis}
To  show the helpfulness of  label enhancement  to {\proposed}, a vanilla variant  of  {\proposed} (named    {\proposedle})  is  adopted.  Here,  label enhancement is replaced by approximating $d_i^j$  with the confidence of the current  model $f_j(\bm{x}_i)$,  which is  a widely-used technique  \cite{feng2020provably,lv2020progressive} to approximate the soft label in weakly supervised learning. 
Table \ref{variants}   reports detailed experimental results in terms of  \emph{Average precision}, \emph{One-error}, and \emph{Ranking loss}, respectively. The detailed experimental results  in  terms  of  other metrics are reported in Appendix F. Besides, the performance of each approach with the number of epochs  on \texttt{scene} is shown in Figure \ref{fig:ablation}.   Wilcoxon signed-ranks test \cite{demvsar2006statistical} in  Table  \ref{Wilcoxon_variants} shows that   {\proposed} achieves superior    performance  against  {\proposedle}  on all evaluation metrics, which clearly validates the usefulness of label enhancement. Figure \ref{fig:convergence} illustrates the estimated  $\mathbf{D}$  converges with the number of epochs  on \texttt{delicious}, which shows that the estimated soft label could  converge efficiently.

\section{Conclusion}
In this paper, we study single-positive multi-label learning and propose    a  novel  approach  {\proposed }.  We  derive an unbiased risk estimator, which  suggests that one positive label of each instance is sufficient to train predictive models for multi-label learning, and  design a benchmark solution via  estimating the soft label   corresponding to  each  example in a  label enhancement process.  The effectiveness of the proposed method is validated  on twelve corrupted MLL datasets. 

\bibliography{xning}

\begin{thebibliography}{10}

\bibitem{bartlett2002rademacher}
Peter~L Bartlett and Shahar Mendelson.
\newblock Rademacher and gaussian complexities: Risk bounds and structural
  results.
\newblock {\em Journal of Machine Learning Research}, 3(Nov):463--482, 2002.

\bibitem{boutell2004learning}
Matthew~R Boutell, Jiebo Luo, Xipeng Shen, and Christopher~M Brown.
\newblock Learning multi-label scene classification.
\newblock {\em Pattern Recognition}, 37(9):1757--1771, 2004.

\bibitem{chen2016deep}
Di~Chen, Yexiang Xue, Shuo Chen, Daniel Fink, and Carla Gomes.
\newblock Deep multi-species embedding.
\newblock {\em arXiv preprint arXiv:1609.09353}, 2016.

\bibitem{chen2019multi}
Zhao-Min Chen, Xiu-Shen Wei, Peng Wang, and Yanwen Guo.
\newblock Multi-label image recognition with graph convolutional networks.
\newblock In {\em Proceedings of the IEEE/CVF Conference on Computer Vision and
  Pattern Recognition}, pages 5177--5186, 2019.

\bibitem{cole2021multi}
Elijah Cole, Oisin Mac~Aodha, Titouan Lorieul, Pietro Perona, Dan Morris, and
  Nebojsa Jojic.
\newblock Multi-label learning from single positive labels.
\newblock In {\em Proceedings of the IEEE/CVF Conference on Computer Vision and
  Pattern Recognition}, pages 933--942, 2021.

\bibitem{demvsar2006statistical}
Janez Dem{\v{s}}ar.
\newblock Statistical comparisons of classifiers over multiple data sets.
\newblock {\em Journal of Machine learning research}, 7(Jan):1--30, 2006.

\bibitem{durand2019learning}
Thibaut Durand, Nazanin Mehrasa, and Greg Mori.
\newblock Learning a deep convnet for multi-label classification with partial
  labels.
\newblock In {\em Proceedings of the IEEE/CVF Conference on Computer Vision and
  Pattern Recognition}, pages 647--657, 2019.

\bibitem{elisseeff2002kernel}
Andr{\'e} Elisseeff and Jason Weston.
\newblock A kernel method for multi-labelled classification.
\newblock In {\em Advances in Neural Information Processing Systems 14 (NIPS
  2002)}, pages 681--687, Vancouver, British Columbia, Canada, 2002.

\bibitem{feng2020provably}
Lei Feng, Jiaqi Lv, Bo~Han, Miao Xu, Gang Niu, Xin Geng, Bo~An, and Masashi
  Sugiyama.
\newblock Provably consistent partial-label learning.
\newblock {\em Advances in Neural Information Processing Systems}, 2020.

\bibitem{figurnov2018implicit}
Michael Figurnov, Shakir Mohamed, and Andriy Mnih.
\newblock Implicit reparameterization gradients.
\newblock {\em Advances in Neural Information Processing Systems}, 2018.

\bibitem{furnkranz2008multilabel}
Johannes F{\"u}rnkranz, Eyke H{\"u}llermeier, Eneldo~Loza Menc{\'\i}a, and
  Klaus Brinker.
\newblock Multilabel classification via calibrated label ranking.
\newblock {\em Machine Learning}, 73(2):133--153, 2008.

\bibitem{GoldbergZRXN10}
Andrew~B. Goldberg, Xiaojin Zhu, Ben Recht, Jun{-}Ming Xu, and Robert~D. Nowak.
\newblock Transduction with matrix completion: Three birds with one stone.
\newblock In {\em Advances in Neural Information Processing Systems 23}, pages
  757--765. Curran Associates, Inc., 2010.

\bibitem{hang2021collaborative}
Jun-Yi Hang and Min-Ling Zhang.
\newblock Collaborative learning of label semantics and deep label-specific
  features for multi-label classification.
\newblock {\em IEEE Transactions on Pattern Analysis and Machine Intelligence},
  2021.

\bibitem{hartvigsen2020recurrent}
Thomas Hartvigsen, Cansu Sen, Xiangnan Kong, and Elke Rundensteiner.
\newblock Recurrent halting chain for early multi-label classification.
\newblock In {\em Proceedings of the 26th ACM SIGKDD International Conference
  on Knowledge Discovery \& Data Mining}, pages 1382--1392, 2020.

\bibitem{huang2016learning}
Jun Huang, Guorong Li, Qingming Huang, and Xindong Wu.
\newblock Learning label-specific features and class-dependent labels for
  multi-label classification.
\newblock {\em IEEE transactions on knowledge and data engineering},
  28(12):3309--3323, 2016.

\bibitem{huynh2020interactive}
Dat Huynh and Ehsan Elhamifar.
\newblock Interactive multi-label cnn learning with partial labels.
\newblock In {\em Proceedings of the IEEE/CVF Conference on Computer Vision and
  Pattern Recognition}, pages 9423--9432, 2020.

\bibitem{KingmaB14}
Diederik~P. Kingma and Jimmy Ba.
\newblock Adam: {A} method for stochastic optimization.
\newblock In {\em 3rd International Conference on Learning Representations,
  2015, San Diego, CA}, 2015.

\bibitem{kingma2014auto}
Diederik~P Kingma and Max Welling.
\newblock Auto-encoding variational bayes.
\newblock In {\em International Conference on Learning Representations}, Banff,
  AB, Canada, 2014.

\bibitem{kipf2016variational}
Thomas~N Kipf and Max Welling.
\newblock Variational graph auto-encoders.
\newblock {\em arXiv preprint arXiv:1611.07308}, 2016.

\bibitem{lanchantin2021general}
Jack Lanchantin, Tianlu Wang, Vicente Ordonez, and Yanjun Qi.
\newblock General multi-label image classification with transformers.
\newblock In {\em Proceedings of the IEEE/CVF Conference on Computer Vision and
  Pattern Recognition}, pages 16478--16488, 2021.

\bibitem{lo2011cost}
Hung-Yi Lo, Ju-Chiang Wang, Hsin-Min Wang, and Shou-De Lin.
\newblock Cost-sensitive multi-label learning for audio tag annotation and
  retrieval.
\newblock {\em IEEE Transactions on Multimedia}, 13(3):518--529, 2011.

\bibitem{lv2020progressive}
Jiaqi Lv, Miao Xu, Lei Feng, Gang Niu, Xin Geng, and Masashi Sugiyama.
\newblock Progressive identification of true labels for partial-label learning.
\newblock In {\em International Conference on Machine Learning}, pages
  6500--6510. PMLR, 2020.

\bibitem{ma2020topic}
Jianghong Ma and Tommy~WS Chow.
\newblock Topic-based instance and feature selection in multilabel
  classification.
\newblock {\em IEEE Transactions on Neural Networks and Learning Systems},
  2020.

\bibitem{ma2021expand}
Zhongchen Ma and Songcan Chen.
\newblock Expand globally, shrink locally: Discriminant multi-label learning
  with missing labels.
\newblock {\em Pattern Recognition}, 111:107675, 2021.

\bibitem{mohri2018foundations}
Mehryar Mohri, Afshin Rostamizadeh, and Ameet Talwalkar.
\newblock {\em Foundations of machine learning}.
\newblock MIT press, 2018.

\bibitem{read2011classifier}
Jesse Read, Bernhard Pfahringer, Geoff Holmes, and Eibe Frank.
\newblock Classifier chains for multi-label classification.
\newblock {\em Machine Learning}, 85(3):333, 2011.

\bibitem{rubin2012statistical}
Timothy~N Rubin, America Chambers, Padhraic Smyth, and Mark Steyvers.
\newblock Statistical topic models for multi-label document classification.
\newblock {\em Machine learning}, 88(1-2):157--208, 2012.

\bibitem{szegedy2016rethinking}
Christian Szegedy, Vincent Vanhoucke, Sergey Ioffe, Jon Shlens, and Zbigniew
  Wojna.
\newblock Rethinking the inception architecture for computer vision.
\newblock In {\em Proceedings of the IEEE conference on computer vision and
  pattern recognition}, pages 2818--2826, 2016.

\bibitem{tang2020multi}
Pingjie Tang, Meng Jiang, Bryan~Ning Xia, Jed~W Pitera, Jeffrey Welser, and
  Nitesh~V Chawla.
\newblock Multi-label patent categorization with non-local attention-based
  graph convolutional network.
\newblock In {\em Proceedings of the AAAI Conference on Artificial
  Intelligence}, volume~34, pages 9024--9031, 2020.

\bibitem{tsoumakas2006multi}
Grigorios Tsoumakas and Ioannis Katakis.
\newblock Multi-label classification: An overview.
\newblock {\em International Journal of Data Warehousing and Mining},
  3(3):1--13, 2006.

\bibitem{tsoumakas2009mining}
Grigorios Tsoumakas, Ioannis Katakis, and Ioannis Vlahavas.
\newblock Mining multi-label data.
\newblock In {\em Data mining and knowledge discovery handbook}, pages
  667--685. Springer, 2009.

\bibitem{tsoumakas2011random}
Grigorios Tsoumakas, Ioannis Katakis, and Ioannis Vlahavas.
\newblock Random k-labelsets for multilabel classification.
\newblock {\em IEEE Transactions on Knowledge and Data Engineering},
  23(7):1079--1089, 2011.

\bibitem{vallet2015multi}
Alexis Vallet and Hiroyasu Sakamoto.
\newblock A multi-label convolutional neural network for automatic image
  annotation.
\newblock {\em Journal of information processing}, 23(6):767--775, 2015.

\bibitem{wu2014multi}
Baoyuan Wu, Zhilei Liu, Shangfei Wang, Bao-Gang Hu, and Qiang Ji.
\newblock Multi-label learning with missing labels.
\newblock In {\em 2014 22nd International Conference on Pattern Recognition},
  pages 1964--1968. IEEE, 2014.

\bibitem{wu2014music}
Bin Wu, Erheng Zhong, Andrew Horner, and Qiang Yang.
\newblock Music emotion recognition by multi-label multi-layer multi-instance
  multi-view learning.
\newblock In {\em Proceedings of the 22nd ACM international conference on
  Multimedia}, pages 117--126, 2014.

\bibitem{xu2019label}
Ning Xu, Yun-Peng Liu, and Xin Geng.
\newblock Label enhancement for label distribution learning.
\newblock {\em IEEE Transactions on Knowledge and Data Engineering}, 33(4):1632
  -- 1643, 2021.

\bibitem{xu2021instance}
Ning Xu, Congyu Qiao, Xin Geng, and Min-Ling Zhang.
\newblock Instance-dependent partial label learning.
\newblock {\em Advances in Neural Information Processing Systems}, 34, 2021.

\bibitem{xu2020variational}
Ning Xu, Jun Shu, Yun-Peng Liu, and Xin Geng.
\newblock Variational label enhancement.
\newblock In {\em Proceedings of the International Conference on Machine
  Learning}, pages 10597--10606, Vienna, Austria, 2020.

\bibitem{Xu2022variational}
Ning Xu, Jun Shu, Ren-Yi Zheng, Xin Geng, Deyu Meng, and Min-Ling Zhang.
\newblock Variational label enhancement.
\newblock {\em IEEE Transactions on Pattern Analysis and Machine Intelligence},
  page in press, 2021.

\bibitem{you2020cross}
Renchun You, Zhiyao Guo, Lei Cui, Xiang Long, Yingze Bao, and Shilei Wen.
\newblock Cross-modality attention with semantic graph embedding for
  multi-label classification.
\newblock In {\em Proceedings of the AAAI Conference on Artificial
  Intelligence}, volume~34, pages 12709--12716, 2020.

\bibitem{yu2014large}
Hsiang-Fu Yu, Prateek Jain, Purushottam Kar, and Inderjit Dhillon.
\newblock Large-scale multi-label learning with missing labels.
\newblock In {\em International conference on machine learning}, pages
  593--601. PMLR, 2014.

\bibitem{yu2021multi}
Ze-Bang Yu and Min-Ling Zhang.
\newblock Multi-label classification with label-specific feature generation: A
  wrapped approach.
\newblock {\em IEEE Transactions on Pattern Analysis and Machine Intelligence},
  2021.

\bibitem{ZhangBHRV17}
Chiyuan Zhang, Samy Bengio, Moritz Hardt, Benjamin Recht, and Oriol Vinyals.
\newblock Understanding deep learning requires rethinking generalization.
\newblock In {\em 5th International Conference on Learning Representations},
  Toulon, France.

\bibitem{zhang2018multi}
Jing Zhang and Xindong Wu.
\newblock Multi-label inference for crowdsourcing.
\newblock In {\em Proceedings of the 24th ACM SIGKDD International Conference
  on Knowledge Discovery \& Data Mining}, pages 2738--2747, 2018.

\bibitem{zhang2007ml}
Min-Ling Zhang and Zhi-Hua Zhou.
\newblock Ml-knn: A lazy learning approach to multi-label learning.
\newblock {\em Pattern Recognition}, 40(7):2038--2048, 2007.

\bibitem{zhang2014review}
Min-Ling Zhang and Zhi-Hua Zhou.
\newblock A review on multi-label learning algorithms.
\newblock {\em IEEE Transactions on Knowledge and Data Engineering},
  26(8):1819--1837, 2014.

\bibitem{zhu2017multi}
Yue Zhu, James~T Kwok, and Zhi-Hua Zhou.
\newblock Multi-label learning with global and local label correlation.
\newblock {\em IEEE Transactions on Knowledge and Data Engineering},
  30(6):1081--1094, 2017.

\end{thebibliography}
\bibliographystyle{plain}

\newpage
\appendix

\section{Calculation Details of Eq. (4)}
\begin{equation}
	\begin{aligned}
		&\sum_{Y\in \mathcal{C}} \mathcal{L}\left( f\left(\bm{x}\right),Y\right)p(Y|\bm{x})\\
		=&\sum_{Y\in \mathcal{C}}\sum_{j \in Y}\ell^{j}p(Y|\bm{x}) + \sum_{Y\in \mathcal{C}}\sum_{j \notin Y}\bar{\ell}^{j}p(Y|\bm{x})\\
		=& \sum_{j=1}^{c}\ell^{j}\sum_{Y\in \mathcal{C}_{j}}p(Y|\bm{x}) + \sum_{j=1}^{c}\bar{\ell}^{j}\sum_{Y\in \widetilde{\mathcal{C}}_j}p(Y|\bm{x})\\
		=& \sum_{j=1}^{c}p(y^j=1|\bm{x})\ell^{j}\sum_{Y\in \mathcal{C}_{j}}\prod_{k\in Y,k\neq j}p(y^k=1|\bm{x})\prod_{k\notin Y}\left( 1 - p(y^k=1|\bm{x})\right) + \\ &\sum_{j=1}^{c}\left(1-p(y^j=1|\bm{x})\right)\bar{\ell}^{j}\sum_{Y\in \widetilde{\mathcal{C}}_j}\prod_{k\in Y}p(y^k=1|\bm{x})\prod_{k\notin Y, k\neq j}\left( 1 - p(y^k=1|\bm{x})\right) \\
		=& \sum_{j=1}^{c}p(y^j=1|\bm{x})\ell^{j} + \left(1-p(y^j=1|\bm{x})\right)\bar{\ell}^{j}\\
		=& \sum_{j=1}^{c}d^j\ell^j + \left(1-d^j\right)\bar{\ell}^j.
	\end{aligned}
\end{equation}
where $d^j=p(y^j=1|\bm{x})$, $\mathcal{C}_j$ denotes the  subset of $\mathcal{C}$  which contains label $j$ and $\widetilde{\mathcal{C}}_j$ denotes the subset of $\mathcal{C}$ without label $j$.

\section{Calculation Details of Eq. (4)}
\begin{equation}
	\log p(\bf L, \bf X, \bf A)=\log p(\bf D, \bf Z, \bf L, \bf X, \bf A)-\log p(\bf D, \bf Z \vert \bf L, \bf X, \bf A).
\end{equation}

Multiply both sides by $q_{\bf w}(\bf Z, \bf D \vert \bf L, \bf X, \bf A)$, and for $\bf D$ and $\bf Z$ integral:
\begin{equation}
	\begin{aligned}
		&\int_{\bf Z, \bf D} q_w(\bf Z, \bf D \mid \bf L, \bf X, \bf A) \log p(\bf L, \bf X, \bf A) d \bf Z d \bf D \\
		&=\int_{\bf Z, \bf D} q_w(\bf Z, \bf D \mid \bf L, \bf X, \bf A)\left(\log p(\bf D, \bf Z, \bf L, \bf X, \bf A)-\log p(\bf D, \bf Z \mid \bf L, \bf X, \bf A)\right) d \bf Z d \bf D.
	\end{aligned}
\end{equation}

On the left side, $\log p(\bf L, \bf X, \bf A)$ is independent of $\bf D$ and $\bf Z$:
\begin{equation}
	\begin{aligned}
		&\log p(\bf L, \bf X, \bf A)=\int_{\bf Z, \bf D} q_{\bf w}(\bf Z, \bf D \mid \bf L, \bf X, \bf A)(\log p(\bf Z, \bf D, \bf L, \bf X, \bf A)\\
		&-\log p(\bf Z, \bf D \mid \bf L, \bf X, \bf A)) d \bf Z d \bf D\\
		&=\int_{\bf Z, \bf D} q_{\bf w}(\bf Z, \bf D \mid \bf L, \bf X, \bf A)\left(\log \frac{p(\bf Z, \bf D, \bf L, \bf X, \bf A)}{q_{\bf w}(\bf Z, \bf D \mid \bf L, \bf X, \bf A)}-\log \frac{p(\bf Z, \bf D \mid \bf L, \bf X, \bf A)}{q_{\bf w}(\bf Z, \bf D \mid \bf L, \bf X, \bf A}\right)\\
		&=\int_{\bf Z, \bf D} q_{\bf w}(\bf Z, \bf D \mid \bf L, \bf X, \bf A) \log \frac{p(\bf Z, \bf D, \bf L, \bf X, \bf A)}{q_{\bf w}(\bf Z, \bf D \mid \bf L, \bf X, \bf A)} d \bf Z d \bf D\\
		&+\text{KL}\left[q_{\bf w}(\bf Z, \bf D \mid \bf L, \bf X, \bf A) \| p(\bf Z, \bf D \mid \bf L, \bf X, \bf A)\right].
	\end{aligned}
\end{equation}

On the right side, the first term is called ELBO:
\begin{equation}
	\mathcal{L}_{ELBO}=\int_{\bf Z, \bf D} q_{\bf w}(\bf Z, \bf D \mid \bf L, \bf X, \bf A) \log \frac{p(\bf Z, \bf D, \bf L, \bf X, \bf A)}{q_{\bf w}(\bf Z, \bf D \mid \bf L, \bf X, \bf A)} d \bf Z d \bf D.
\end{equation}

Then we have:
\begin{equation}
	\log p(\bf L, \bf X, \bf A)=\mathcal{L}_{ELBO} + \text{KL}\left[q_{\bf w}(\bf Z, \bf D \mid \bf L, \bf X, \bf A) \| p(\bf Z, \bf D \mid \bf L, \bf X, \bf A)\right].
\end{equation}

$\mathcal{L}_{ELBO}$ can be calculated as:
\begin{equation}
	\begin{aligned}
		\mathcal{L}_{ELBO}&=\int_{\bf Z, \bf D} q_{\bf w}(\bf Z, \bf D \mid \bf L, \bf X, \bf A) \log \frac{p(\bf Z, \bf D, \bf L, \bf X, \bf A)}{q_{\bf w}(\bf Z, \bf D \mid \bf L, \bf X, \bf A)} d \bf Z d \bf D\\
		&=\mathbb E_{q_{\bf w}(\bf Z, \bf D \mid \bf L, \bf X, \bf A)}\left[\log \frac{p(\bf Z, \bf D, \bf L, \bf X, \bf A)}{q_{\bf w}(\bf Z, \bf D \mid \bf L, \bf X, \bf A)}\right] \\
		&=\mathbb E_{q_{\bf w}(\bf Z, \bf D \mid \bf L, \bf X, \bf A)}\left[\log \frac{p(\bf Z) p(\bf D) p(\bf L, \bf X, \bf A \mid \bf Z, \bf D)}{q_{{\bf w}_1}(\bf D \mid \bf L, \bf X, \bf A) q_{{\bf w}_2}(\bf Z \mid \bf D, \bf X)}\right] \\
		&=\mathbb E_{q_{\bf w}(\bf Z, \bf D \mid \bf L, \bf X, \bf A)[\log p(\bf L, \bf X, \bf A \mid \bf Z, \bf D)]}\\
		&+\mathbb E_{q_{\bf w}(\bf Z, \bf D \mid \bf L, \bf X, \bf A)}\left[\log \frac{p(\bf Z) p(\bf D)}{q_{{\bf w}_ 1}(\bf D \mid \bf L, \bf X, \bf A) q_{{\bf w}_2}(\bf Z \mid \bf D, \bf X)}\right].
	\end{aligned}
\end{equation}

The first term of $\mathcal{L}_{ELBO}$ can be calculated as:
\begin{equation}
	\begin{aligned}
		\mathbb E_{q_{\bf w}(\bf Z, \bf D \mid \bf L, \bf X, \bf A)[\log p(\bf L, \bf X, \bf A \mid \bf Z, \bf D)]}
		&=\mathbb E_{q_{\bf w}(\bf Z, \bf D \mid \bf L, \bf X, \bf A)}[\log p(\bf X \mid \bf Z, \bf D)] \\
		&+\mathbb E_{q_{\bf w}\left(\bf Z, \bf D \mid \bf L, \bf X, \bf A\right)}[\log P(\bf L \mid \bf D)] \\
		&+\mathbb E_{q_{\bf w}(\bf Z, \bf D \mid \bf L, \bf X, \bf A)}[\log p(\bf A \mid \bf D)].
	\end{aligned}
\end{equation}

The second term of $\mathcal{L}_{ELBO}$ can be calculated as:
\begin{equation}
	\begin{aligned}
		&\mathbb E_{q_{\bf w}}(\bf Z, \bf D \mid \bf L, \bf X, \bf A)[\log p(\bf A \mid \bf D)]\\
		&=\mathbb E_{q_{{\bf w}_1(\bf D \mid \bf L, \bf X, \bf A)}} \mathbb E_{q_{{\bf w}_2}}(\bf Z \mid \bf D, \bf X)\left[\log \frac{p(\bf Z) p(\bf D)}{q_{{\bf w}_1}(\bf D \mid \bf L, \bf X, \bf A) q_{{\bf w}_2}(\bf Z \mid \bf D, \bf X)}\right] \\
		&= \mathbb E_{q_{{\bf w}_1}(\bf D \mid \bf L, \bf X, \bf A)} \mathbb E_{q_{{\bf w}_2}(\bf Z \mid \bf D, \bf X)}\left[\log \frac{p(\bf D)}{q_{{\bf w}_1}(\bf D \mid \bf L, \bf X, \bf A)}\right] \\
		&+ \mathbb E_{q_{{\bf w}_1}(\bf D \mid \bf L, \bf X, \bf A)} \mathbb E_{{q}_{{\bf w}_2}(\bf Z \mid \bf D, \bf X)}\left[\log \frac{p(\bf Z)}{q_{{\bf w}_2}(\bf Z \mid \bf D, \bf X)}\right] \\
		& =-\text{KL}\left[q_{{\bf w}_1}(\bf D \mid \bf L, \bf X, \bf A) \| p(\bf D)\right]
		-\text{KL}\left[q_{{\bf w}_2}(\bf Z \mid \bf D, \bf X) \| p(\bf Z)\right].
	\end{aligned}
\end{equation}

Then we have:
\begin{equation}
	\begin{aligned}
		\mathcal{L}_{ELBO}&=\mathbb E_{q_{\bf w}(\bf Z, \bf D \mid \bf L, \bf X, \bf A)}[\log p(\bf X \mid \bf Z, \bf D)+\log P(\bf L \mid \bf D)+\log p(\bf A \mid \bf D)] \\
		&-\text{KL}\left[q_{{\bf w}_1}(\bf D \mid \bf L, \bf X, \bf A) \| p(\bf D)\right]
		-\text{KL}\left[q_{{\bf w}_2}(\bf Z \mid \bf D, \bf X) \| p(\bf Z)\right].
	\end{aligned}
\end{equation}

\section{Proof of Lemma 1}
In order to prove this lemma, we first show that the one direction $\sup_{f\in\mathcal{F}}R_{sp}(f)-\widehat{R}_{sp}(f)$ is bounded with probability at least $1 - {\delta}/{2}$, and the other direction can be similarly shown. Suppose an example $(\bm{x}, y)$ is replaced by another arbitrary example  $(\bm{x}', y')$, then the change of $\sup_{f\in\mathcal{F}}R_{sp}(f)-\widehat{R}_{sp}(f)$ is no greater than ${M}/{(2n)}$, the loss function $\mathcal{L}_{sp}$ are bounded by $M$. By applying McDiarmid's inequality, for any $\delta \textgreater 0$, with probability at least $1 - \delta/2$,
\begin{equation}
	\sup_{f\in\mathcal{F}}R_{sp}(f)-\widehat{R}_{sp}(f) \leq \mathbb{E}\left[\sup_{f\in\mathcal{F}} R_{sp}(f)-\widehat{R}_{sp}(f) \right] + \frac{M}{2} \sqrt{\frac{\log\frac{2}{\delta}}{2n}}.
\end{equation} 

By sysmmetrization, we can obtain
\begin{equation}
	\mathbb{E}\left[\sup_{f\in\mathcal{F}} R_{sp}(f)-\widehat{R}_{sp}(f) \right] \leq 2\widetilde{\mathfrak{R}}_n(\mathcal{G}_{sp}).
\end{equation}
By further taking into account the other side $\sup_{f\in\mathcal{F}} R_{sp}(f)-\widehat{R}_{sp}(f)$, we have for any $\delta \textgreater 0$, with probability at least $1 - \delta$,
\begin{equation}
	\sup_{f\in\mathcal{F}}\abs{R_{sp}(f)-\widehat{R}_{sp}(f)}\leq 2\widetilde{\mathfrak{R}}_n(\mathcal{G}_{sp})
	+ \frac{M}{2} \sqrt{\frac{\log\frac{2}{\delta}}{2n}}.
\end{equation}

\begin{table*}[t]
	\normalsize
	\centering
	\caption{Characteristics of the experimental datasets.}
	\label{datasets}
	\begin{tabular}{lcccccc}
		\hline
		\hline 
		Dataset & $|\mathcal{S}|$ & $\operatorname{dim}(\mathcal{S})$ & $L(\mathcal{S})$   & Domain \\
		\hline CAL500 & 502 & 68 & 174    & Music\\
		image & 2000 & 294 & 5    & Images  \\
		scene & 2407 & 294 & 6   & Images \\
		yeast & 2417 & 103 & 14   & Biology \\
		corel5k & 5000 & 499 & 374    & Images   \\
		rcv1-s1 & 6000 & 944 & 101    & Text   \\
		corel16k-s1 & 13766 & 500 & 153   & Images  \\
		delicious & 16105 & 500 & 983   & Text \\
		iaprtc12 & 19627 & 1000 & 291    & Images   \\
		espgame & 20770 & 1000 & 268    & Images   \\
		mirflickr & 25000 & 1000 & 38    & Images   \\
		tmc2007 & 28596 & 981 & 22   & Text  \\
		\hline
		\hline
	\end{tabular}
\end{table*}

\begin{table*}[t]
	\centering
	\fontsize{6}{7}\selectfont
	\caption{Predictive performance of each comparing approach (mean$\pm$std) in terms of  \emph{Hamming loss} $\downarrow$.  The best performance (the smaller the better) is shown in bold face.}\label{Hamming}
	\begin{tabular}{ccccccccc}
		\toprule
		Datasets &{\proposed} &{\an} & {\anls} & {\wan} & {\role} & {\glocal} & {\mlml} & {\dml}  \\
		\midrule
		CAL500 & \textbf{0.148$\pm$0.000} & 0.148$\pm$0.000 & 0.149$\pm$0.001 & 0.296$\pm$0.007 & 0.148$\pm$0.000 & 0.148$\pm$0.000 & 0.148$\pm$0.000 & 0.148$\pm$0.000 \\ 
		image & \textbf{0.205$\pm$0.008} & 0.216$\pm$0.012 & 0.213$\pm$0.014 & 0.321$\pm$0.050 & 0.214$\pm$0.019 & 0.211$\pm$0.004 & 0.227$\pm$0.005 & 0.712$\pm$0.018 \\ 
		scene & \textbf{0.124$\pm$0.035} & 0.141$\pm$0.021 & 0.137$\pm$0.023 & 0.193$\pm$0.029 & 0.174$\pm$0.014 & 0.149$\pm$0.017 & 0.174$\pm$0.019 & 0.288$\pm$0.007 \\ 
		yeast & \textbf{0.205$\pm$0.003} & 0.306$\pm$0.000 & 0.306$\pm$0.000 & 0.215$\pm$0.003 & 0.213$\pm$0.006 & 0.277$\pm$0.073 & 0.306$\pm$0.035 & 0.694$\pm$0.015 \\ 
		corel5k & \textbf{0.010$\pm$0.000} & 0.010$\pm$0.000 & 0.010$\pm$0.000 & 0.038$\pm$0.002 & 0.010$\pm$0.000 & 0.010$\pm$0.000 & 0.010$\pm$0.000 & 0.020$\pm$0.000 \\ 
		rcv1-s1 & \textbf{0.027$\pm$0.000} & 0.028$\pm$0.000 & 0.028$\pm$0.000 & 0.047$\pm$0.004 & 0.028$\pm$0.000 & 0.029$\pm$0.000 & 0.029$\pm$0.000 & 0.917$\pm$0.000 \\ 
		corel16k-s1 & \textbf{0.019$\pm$0.004} & 0.019$\pm$0.000 & 0.019$\pm$0.000 & 0.136$\pm$0.005 & 0.019$\pm$0.000 & 0.019$\pm$0.000 & 0.019$\pm$0.000 & 0.077$\pm$0.000 \\ 
		delicious & \textbf{0.019$\pm$0.001} & 0.019$\pm$0.000 & 0.019$\pm$0.000 & 0.075$\pm$0.007 & 0.019$\pm$0.000 & 0.019$\pm$0.000 & 0.019$\pm$0.000 & 0.326$\pm$0.000 \\ 
		iaprtc12 & \textbf{0.019$\pm$0.011} & 0.019$\pm$0.000 & 0.019$\pm$0.000 & 0.195$\pm$0.007 & 0.019$\pm$0.000 & 0.019$\pm$0.000 & 0.019$\pm$0.000 & 0.019$\pm$0.000 \\ 
		espgame & \textbf{0.017$\pm$0.003} & 0.017$\pm$0.000 & 0.017$\pm$0.000 & 0.174$\pm$0.009 & 0.017$\pm$0.000 & 0.017$\pm$0.000 & 0.017$\pm$0.000 & 0.017$\pm$0.000 \\ 
		mirflickr & \textbf{0.118$\pm$0.001} & 0.127$\pm$0.000 & 0.127$\pm$0.000 & 0.211$\pm$0.003 & 0.130$\pm$0.005 & 0.128$\pm$0.000 & 0.128$\pm$0.000 & 0.128$\pm$0.000 \\ 
		tmc2007 & \textbf{0.063$\pm$0.000} & 0.085$\pm$0.001 & 0.089$\pm$0.001 & 0.092$\pm$0.004 & 0.065$\pm$0.002 & 0.098$\pm$0.001 & 0.098$\pm$0.001 & 0.098$\pm$0.000 \\ 
		\bottomrule
	\end{tabular}
\end{table*}

\begin{table*}[t]
	\centering
	\fontsize{6}{7}\selectfont
	\caption{Predictive performance of each comparing approach (mean$\pm$std) in terms of  \emph{Ranking loss} $\downarrow$.  The best performance (the smaller the better) is shown in bold face.}
	\begin{tabular}{ccccccccc}
		\toprule
		Datasets &{\proposed} &{\an} & {\anls} & {\wan} & {\role} & {\glocal} & {\mlml} & {\dml}  \\
		\midrule
		CAL500 & \textbf{0.239$\pm$0.010} & 0.266$\pm$0.045 & 0.391$\pm$0.048 & 0.244$\pm$0.005 & 0.384$\pm$0.010 & 0.366$\pm$0.009 & 0.478$\pm$0.001 & 0.506$\pm$0.013 \\ 
		image & 0.170$\pm$0.055 & 0.330$\pm$0.092 & 0.325$\pm$0.084 & 0.240$\pm$0.045 & 0.234$\pm$0.034 & 0.179$\pm$0.004 & \textbf{0.163$\pm$0.003} & 0.459$\pm$0.014 \\ 
		scene & 0.086$\pm$0.045 & 0.170$\pm$0.132 & 0.171$\pm$0.119 & 0.108$\pm$0.014 & 0.163$\pm$0.045 & 0.108$\pm$0.006 & \textbf{0.056$\pm$0.007} & 0.383$\pm$0.035 \\ 
		yeast & \textbf{0.161$\pm$0.003} & 0.165$\pm$0.002 & 0.168$\pm$0.002 & 0.163$\pm$0.001 & 0.168$\pm$0.001 & 0.332$\pm$0.007 & 0.361$\pm$0.000 & 0.488$\pm$0.007 \\ 
		corel5k & 0.134$\pm$0.003 & 0.113$\pm$0.001 & 0.189$\pm$0.011 & \textbf{0.111$\pm$0.001} & 0.266$\pm$0.013 & 0.139$\pm$0.002 & 0.355$\pm$0.003 & 0.484$\pm$0.001 \\ 
		rcv1-s1 & \textbf{0.042$\pm$0.000} & 0.046$\pm$0.001 & 0.060$\pm$0.001 & 0.042$\pm$0.000 & 0.071$\pm$0.004 & 0.168$\pm$0.003 & 0.179$\pm$0.007 & 0.437$\pm$0.002 \\ 
		corel16k-s1 & \textbf{0.133$\pm$0.001} & 0.138$\pm$0.002 & 0.181$\pm$0.002 & 0.134$\pm$0.001 & 0.241$\pm$0.006 & 0.690$\pm$0.001 & 0.306$\pm$0.005 & 0.454$\pm$0.002 \\ 
		delicious & 0.126$\pm$0.000 & 0.133$\pm$0.002 & 0.276$\pm$0.015 & \textbf{0.125$\pm$0.001} & 0.306$\pm$0.007 & 0.445$\pm$0.011 & 0.325$\pm$0.004 & 0.456$\pm$0.004 \\ 
		iaprtc12 & \textbf{0.115$\pm$0.002} & 0.128$\pm$0.003 & 0.230$\pm$0.011 & 0.140$\pm$0.005 & 0.167$\pm$0.002 & 0.442$\pm$0.003 & 0.266$\pm$0.011 & 0.502$\pm$0.015 \\ 
		espgame & \textbf{0.158$\pm$0.001} & 0.163$\pm$0.006 & 0.268$\pm$0.004 & 0.158$\pm$0.001 & 0.241$\pm$0.006 & 0.464$\pm$0.001 & 0.319$\pm$0.023 & 0.500$\pm$0.003 \\ 
		mirflickr & \textbf{0.117$\pm$0.002} & 0.118$\pm$0.001 & 0.148$\pm$0.003 & 0.123$\pm$0.002 & 0.155$\pm$0.006 & 0.189$\pm$0.019 & 0.944$\pm$0.003 & 0.496$\pm$0.007 \\ 
		tmc2007 & 0.049$\pm$0.001 & 0.047$\pm$0.001 & 0.060$\pm$0.002 & \textbf{0.045$\pm$0.001} & 0.061$\pm$0.002 & 0.144$\pm$0.003 & 0.143$\pm$0.001 & 0.453$\pm$0.001 \\ 
		\bottomrule
	\end{tabular}
	\label{ranking}
\end{table*}

\begin{table*}[htb]
	\centering
	\fontsize{6}{7}\selectfont
	\caption{Predictive performance of each comparing approach (mean$\pm$std) in terms of  \emph{Coverage} $\downarrow$.  The best performance (the smaller the better) is shown in bold face.}\label{Coverage}
	\begin{tabular}{ccccccccc}
		\toprule
		Datasets &{\proposed} &{\an} & {\anls} & {\wan} & {\role} & {\glocal} & {\mlml} & {\dml}  \\
		\midrule
		CAL500 & 0.865$\pm$0.008 & 0.881$\pm$0.014 & 0.937$\pm$0.017 & 0.878$\pm$0.015 & 0.953$\pm$0.012 & 0.875$\pm$0.013 & \textbf{0.668$\pm$0.001} & 0.694$\pm$0.003 \\ 
		image & \textbf{0.171$\pm$0.045} & 0.298$\pm$0.075 & 0.294$\pm$0.069 & 0.225$\pm$0.037 & 0.221$\pm$0.028 & 0.177$\pm$0.018 & 0.783$\pm$0.005 & 0.966$\pm$0.014 \\ 
		scene & \textbf{0.084$\pm$0.037} & 0.155$\pm$0.112 & 0.156$\pm$0.101 & 0.102$\pm$0.012 & 0.146$\pm$0.036 & 0.103$\pm$0.002 & 0.414$\pm$0.002 & 0.931$\pm$0.004 \\ 
		yeast & \textbf{0.455$\pm$0.007} & 0.456$\pm$0.008 & 0.469$\pm$0.010 & 0.460$\pm$0.004 & 0.476$\pm$0.004 & 0.689$\pm$0.001 & 0.942$\pm$0.003 & 0.951$\pm$0.002 \\ 
		corel5k & 0.312$\pm$0.007 & \textbf{0.273$\pm$0.002} & 0.447$\pm$0.022 & 0.273$\pm$0.001 & 0.557$\pm$0.025 & 0.328$\pm$0.005 & 0.396$\pm$0.008 & 0.465$\pm$0.016 \\ 
		rcv1-s1 & \textbf{0.107$\pm$0.001} & 0.117$\pm$0.003 & 0.153$\pm$0.004 & 0.107$\pm$0.000 & 0.177$\pm$0.007 & 0.315$\pm$0.004 & 0.439$\pm$0.002 & 0.731$\pm$0.003 \\ 
		corel16k-s1 & \textbf{0.269$\pm$0.003} & 0.280$\pm$0.006 & 0.364$\pm$0.005 & 0.271$\pm$0.001 & 0.465$\pm$0.010 & 0.847$\pm$0.007 & 0.740$\pm$0.004 & 0.848$\pm$0.006 \\ 
		delicious & 0.630$\pm$0.002 & 0.647$\pm$0.012 & 0.894$\pm$0.013 & \textbf{0.626$\pm$0.003} & 0.910$\pm$0.004 & 0.861$\pm$0.009 & 0.749$\pm$0.019 & 0.829$\pm$0.002 \\ 
		iaprtc12 & \textbf{0.336$\pm$0.003} & 0.361$\pm$0.007 & 0.593$\pm$0.019 & 0.377$\pm$0.010 & 0.446$\pm$0.005 & 0.695$\pm$0.011 & 0.793$\pm$0.007 & 0.934$\pm$0.008 \\ 
		espgame & \textbf{0.382$\pm$0.004} & 0.395$\pm$0.017 & 0.603$\pm$0.009 & 0.384$\pm$0.002 & 0.556$\pm$0.012 & 0.721$\pm$0.018 & 0.850$\pm$0.004 & 0.935$\pm$0.006 \\ 
		mirflickr & \textbf{0.327$\pm$0.003} & 0.328$\pm$0.003 & 0.397$\pm$0.005 & 0.332$\pm$0.002 & 0.396$\pm$0.013 & 0.436$\pm$0.016 & 0.944$\pm$0.012 & 0.990$\pm$0.003 \\ 
		tmc2007 & 0.130$\pm$0.002 & 0.124$\pm$0.002 & 0.149$\pm$0.004 & \textbf{0.120$\pm$0.001} & 0.150$\pm$0.004 & 0.264$\pm$0.004 & 0.834$\pm$0.003 & 0.985$\pm$0.000 \\ 
		\bottomrule
	\end{tabular}
\end{table*}

\begin{table*}[t]
	\centering
	\caption{Predictive performance of {\proposed} and its variant (mean$\pm$std) in terms of  \emph{Hamming Loss} and  \emph{Coverage}.}
	\begin{tabular}{ccccc}
		\hline
		\multirow{2}*{Datasets} & \multicolumn{2}{c}{\emph{Hamming loss} $\downarrow$ } & \multicolumn{2}{c}{\emph{Coverage} $\downarrow$ } \\ 
		
		\cline{2-5}
		~ & {\proposed} & {\proposedle} & {\proposed} & {\proposedle} \\
		\hline
		CAL500 & \textbf{0.148$\pm$0.000} & 0.148$\pm$0.000 & \textbf{0.865$\pm$0.008} & 0.897$\pm$0.002 \\ 
		image & \textbf{0.205$\pm$0.008} & 0.229$\pm$0.000 & \textbf{0.171$\pm$0.045} & 0.376$\pm$0.007 \\
		scene & \textbf{0.124$\pm$0.035} & 0.169$\pm$0.008 & \textbf{0.084$\pm$0.037} & 0.152$\pm$0.030 \\
		yeast & \textbf{0.205$\pm$0.003} & 0.306$\pm$0.000 & \textbf{0.455$\pm$0.007} & 0.457$\pm$0.003 \\
		corel5k & \textbf{0.010$\pm$0.000} & 0.010$\pm$0.000 & 0.312$\pm$0.007 & \textbf{0.282$\pm$0.001} \\
		rcv1-s1 & \textbf{0.027$\pm$0.000} & 0.029$\pm$0.000 & \textbf{0.107$\pm$0.001} & 0.138$\pm$0.001 \\
		corel16k-s1 & \textbf{0.019$\pm$0.004} & 0.019$\pm$0.000 & \textbf{0.269$\pm$0.003} & 0.283$\pm$0.000 \\
		delicious & \textbf{0.019$\pm$0.001} & 0.019$\pm$0.000 & \textbf{0.630$\pm$0.002} & 0.663$\pm$0.005 \\
		iaprtc12 & \textbf{0.019$\pm$0.011} & 0.019$\pm$0.000 & \textbf{0.336$\pm$0.003} & 0.403$\pm$0.000 \\
		espgame & \textbf{0.017$\pm$0.003} & 0.017$\pm$0.000 & \textbf{0.382$\pm$0.004} & 0.412$\pm$0.005 \\
		mirflickr & \textbf{0.118$\pm$0.001} & 0.128$\pm$0.000 & \textbf{0.327$\pm$0.003} & 0.335$\pm$0.002 \\
		tmc2007 & \textbf{0.063$\pm$0.000} & 0.098$\pm$0.000 & 0.130$\pm$0.002 & \textbf{0.127$\pm$0.000} \\
		\bottomrule
	\end{tabular}
	\label{Ablation_2}
\end{table*}

\section{Proof of Lemma 2}
As $w^j$ and $\bar{w}^j$ are bounded in $[0, \kappa]$, we can obtain $\widetilde{\mathfrak{R}}_n(\mathcal{G}_{spl}) \leq \kappa c\left(\mathfrak{R}_n(\ell \circ \mathcal{F})+ \mathfrak{R}_n(\bar{\ell} \circ \mathcal{F})\right)$ where $\ell \circ \mathcal{F}$ denotes $\{\ell \circ \mathcal{F} | f\in \mathcal{F}\}$ and $\bar{\ell} \circ \mathcal{F}$ denotes $\{\bar{\ell} \circ \mathcal{F} | f\in \mathcal{F}\}$. Since $\mathcal{H}_{y} = \{ h: \bm{x} \mapsto f_{y}(\bm{x})|f\in \mathcal{F}\}$ and the loss functions $\ell\left(f(\bm{x}),{y}\right)$ and $\bar{\ell}\left(f(\bm{x}),{y}\right)$ are $\rho^{+}$-Lipschitz and $\rho^{-}$-Lipschitz with respect to $f(\bm{x})$ $(0\textless \rho^{+} \textless \infty$ and $0\textless \rho^{-} \textless \infty )$ for all $y \in \mathcal{Y}$, by the Rademacher vector contraction inequality, we have $\mathfrak{R}_n(\ell \circ \mathcal{F}) + \mathfrak{R}_n(\bar{\ell} \circ \mathcal{F})\leq \sqrt{2}(\rho^{+}+\rho^{-})\sum_{j=1}^{c}\mathfrak{R}_n(\mathcal{H}_{y})$.

\section{Proof of Theorem 1}
Combining Lemma 1 and 2, we have
\begin{equation}
	\begin{split}
		R(\widehat{f}_{sp}) - R(f^{*}) &= R(\widehat{f}_{sp}) - \widehat{R}_{sp}(\widehat{f}) + \widehat{R}_{sp}(\widehat{f}) - \widehat{R}_{sp}(f^{*}) + \widehat{R}_{sp}(f^{*}) - R(f^{*}) \\
		&\leq R(\widehat{f}_{sp}) - \widehat{R}_{sp}(\widehat{f}) + \widehat{R}_{sp}(f^{*}) - R(f^{*}) \\ 
		&\leq 2\sup_{f\in\mathcal{F}}\abs{R_{sp}(f)-\widehat{R}_{sp}(f)} \\
		&\leq 4\widetilde{\mathfrak{R}}_n(\mathcal{G}_{sp}) + M \sqrt{\frac{\log\frac{2}{\delta}}{2n}} \\
		&\leq 4\sqrt{2}\kappa c(\rho^{+} + \rho^{-})\sum_{j=1}^{c}\mathfrak{R}_n(\mathcal{H}_{y}) + M \sqrt{\frac{\log\frac{2}{\delta}}{2n}}.
	\end{split}
\end{equation}
which concludes the proof.
\section{Details of Experiments}
Some basic statistics about these  datasets are given in Table \ref{datasets}, including the number of examples $(|S|)$, the number of features $(\operatorname{dim}(S))$, and the number of class labels $(L(S))$. Tables \ref{Hamming} to  \ref{Coverage}  show the   results of all   approaches  on \emph{One-error}, \emph{Hamming loss}, and \emph{Coverage}, respectively.
Tables \ref{Ablation_2} shows the   results of {\proposed} and its variant {\proposedle} (mean$\pm$std) in terms of  \emph{Hamming Loss} and  \emph{Coverage}.

\end{document}